\pdfoutput=1

\documentclass[11pt]{article}

\usepackage[]{emnlp2021}

\usepackage{times}
\usepackage{latexsym}

\usepackage[T1]{fontenc}

\usepackage[utf8]{inputenc}

\usepackage{microtype}

\usepackage{graphicx}
\graphicspath{ {./figures/} }

\usepackage{multirow}

\usepackage{bbm}

\usepackage{amsfonts,amssymb}

\usepackage[normalem]{ulem}
\useunder{\uline}{\ul}{}

\title{A More Fine-Grained Aspect-Sentiment-Opinion Triplet Extraction Task}

\author{
	Yuncong Li \renewcommand{\thefootnote}{\arabic{footnote}}\footnotemark[1], Fang Wang \renewcommand{\thefootnote}{\arabic{footnote}}\footnotemark[2],  
	Wenjun Zhang \renewcommand{\thefootnote}{\arabic{footnote}}\footnotemark[3], 
	Sheng-hua Zhong \renewcommand{\thefootnote}{\arabic{footnote}}\footnotemark[2] \renewcommand{\thefootnote}{\fnsymbol{footnote}}\footnotemark[2], \\ 
	\textbf{Cunxiang Yin} \renewcommand{\thefootnote}{\arabic{footnote}}\footnotemark[1], and \textbf{Yancheng He} \renewcommand{\thefootnote}{\arabic{footnote}}\footnotemark[1] \\
	
	\renewcommand{\thefootnote}{\arabic{footnote}}\footnotemark[1] Tencent Inc. \\
	\texttt{\{yuncongli,jasonyin,collinhe\}@tencent.com} \\
	\renewcommand{\thefootnote}{\arabic{footnote}}\footnotemark[2] College of Computer Science and Software Engineering, Shenzhen University, \\ Shenzhen, China \\
	\texttt{2160230414@email.szu.edu.cn,csshzhong@szu.edu.cn} \\
	\renewcommand{\thefootnote}{\arabic{footnote}}\footnotemark[3] Baidu Inc.\\
}

\begin{document}
\maketitle

\renewcommand{\thefootnote}{\fnsymbol{footnote}}
\footnotetext[2]{Corresponding author}
\renewcommand{\thefootnote}{\arabic{footnote}}

\begin{abstract}
Aspect Sentiment Triplet Extraction (ASTE) aims to extract aspect term, sentiment and opinion term triplets from sentences and tries to provide a complete solution for aspect-based sentiment analysis (ABSA). However, some triplets extracted by ASTE are confusing, since the sentiment in a triplet extracted by ASTE is the sentiment that the sentence expresses toward the aspect term rather than the sentiment of the aspect term and opinion term pair. In this paper, we introduce a more fine-grained Aspect-Sentiment-Opinion Triplet Extraction (ASOTE) Task. ASOTE also extracts aspect term, sentiment and opinion term triplets. However, the sentiment in a triplet extracted by ASOTE is the sentiment of the aspect term and opinion term pair. We build four datasets for ASOTE based on several popular ABSA benchmarks. We propose a Position-aware BERT-based Framework (PBF) to address this task. PBF first extracts aspect terms from sentences. For each extracted aspect term, PBF first generates aspect term-specific sentence representations considering both the meaning and the position of the aspect term, then extracts associated opinion terms and predicts the sentiments of the aspect term and opinion term pairs based on the sentence representations. Experimental results on the four datasets show the effectiveness of PBF. \footnote{Data and code are available at https://github.com/l294265421/ASOTE}
\end{abstract}

\section{Introduction}
\label{sec:introduction}

\begin{figure}
	\centering
	\includegraphics[scale=0.15]{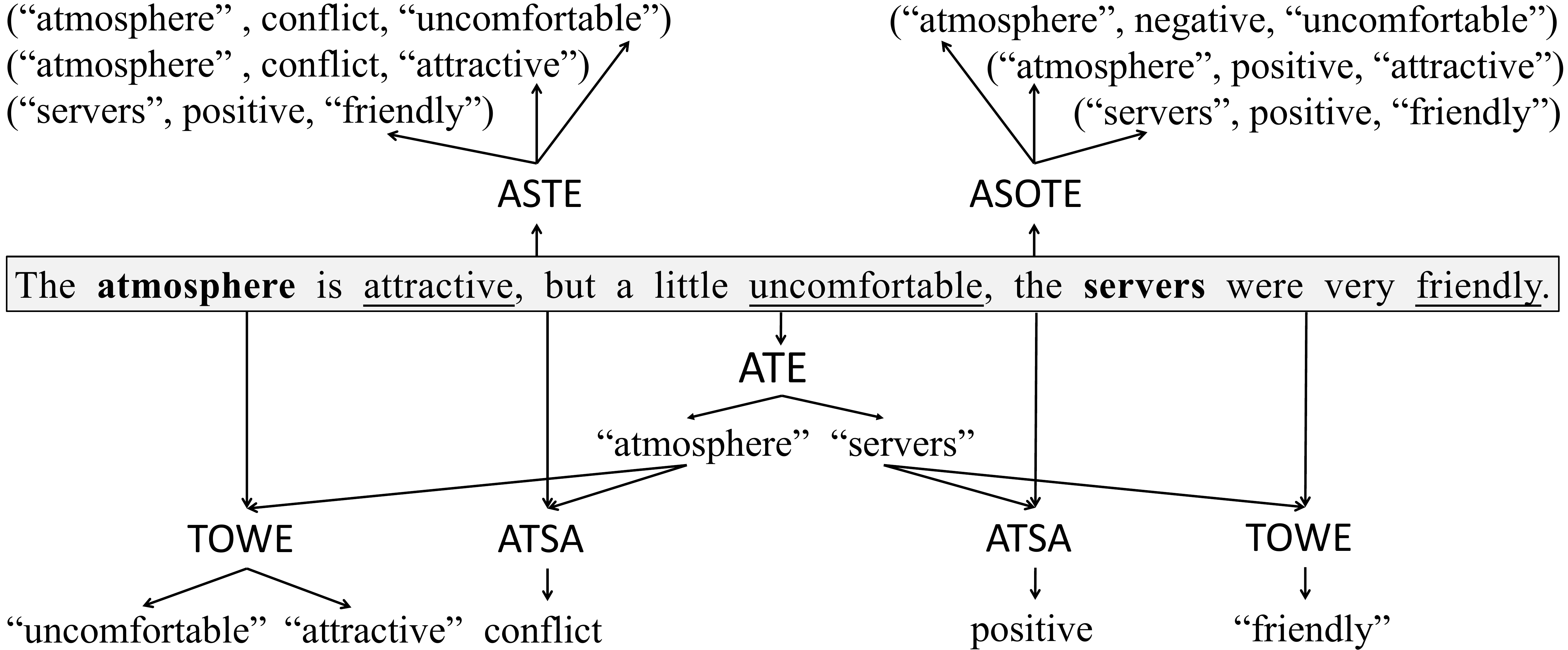}
	\caption{An example showing the inputs and outputs of the tasks. For each arrow, when the head is a task name, the tail is an input of the task; when the tail is a task name, the head is an output of the task. The bold words are aspects. The underlined words are opinions.}
	\label{fig:tasks}
\end{figure}

Aspect-based sentiment analysis (\textbf{ABSA}) \citep{10.1145/1014052.1014073, pontiki-etal-2014-semeval, pontiki-etal-2015-semeval, pontiki-etal-2016-semeval} is a fine-grained sentiment analysis~\citep{10.1145/945645.945658, liu2012sentiment} task and can provide more detailed information than general sentiment analysis. To solve the ABSA task, many subtasks have been proposed, such as, Aspect Term Extraction (\textbf{ATE}), Aspect Term Sentiment Analysis (\textbf{ATSA}) and Target-oriented Opinion Words Extraction (\textbf{TOWE}) \citep{fan2019target}. An \textbf{aspect term} (\textbf{aspect} for short) is a word or phrase that refers to a discussed entity in a sentence. An \textbf{opinion term} (\textbf{opinion} for short) is a word or phrase that expresses a subjective attitude. ATE extracts aspects from sentences. Given a sentence and an aspect in the sentence, ATSA and TOWE predict the sentiment and opinions associated with the aspect. These subtasks can work together to tell a complete story, i.e. the discussed aspect, the sentiment of the aspect, and the cause of the sentiment. However, no previous ABSA study tried to provide a complete solution in one shot. \citet{Peng2020KnowingWH} proposed the Aspect Sentiment Triplet Extraction (\textbf{ASTE}) task, which attempted to provide a complete solution for ABSA. A triplet extracted from a sentence by ASTE contains an aspect, the sentiment that the sentence expresses toward the aspect, and one opinion associated with the aspect. The example in Figure~\ref{fig:tasks} shows the inputs and outputs of the tasks mentioned above.

\begin{figure*}
	\centering
	\includegraphics[scale=0.4]{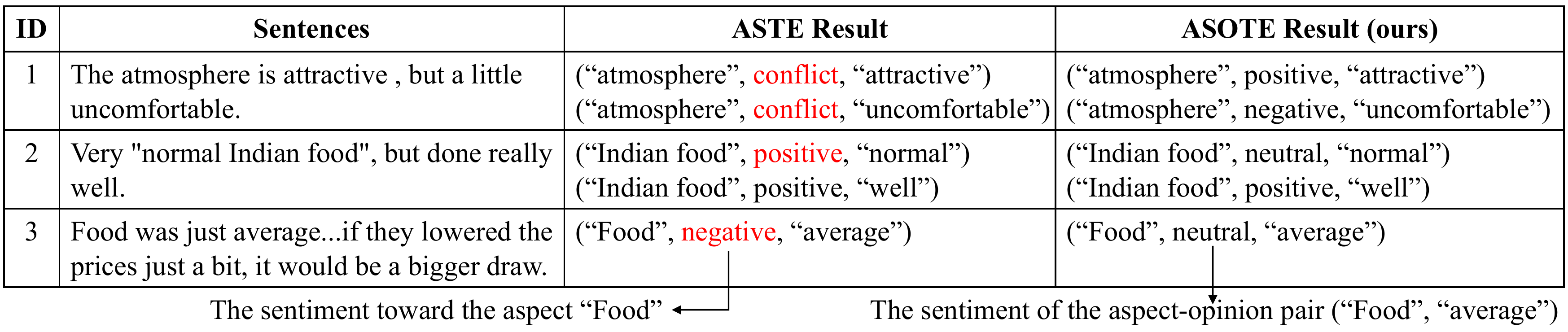}
	\caption{Differences between ASOTE and ASTE. In the third sentence, the negative sentiment toward the aspect ``Food'' is expressed without an annotatable opinion.}
	\label{fig:asote_aste}
\end{figure*}

However, the triplet extracted from a sentence by ASTE becomes confusing when the sentence has multiple opinions about the aspect and these opinions express different sentiments toward the aspect, since the sentiment in a triplet extracted by ASTE is the sentiment that the sentence expresses toward the aspect rather than the sentiment of the aspect and opinion pair. The third column in Figure~\ref{fig:asote_aste}  shows the extraction results of ASTE from the corresponding sentences. When seeing the triplets where the words indicating sentiments are red, people will be confused. Moreover, downstream tasks can not benefit from these triplets. 

In this paper,  we introduce a more fine-grained Aspect-Sentiment-Opinion Triplet Extraction (\textbf{ASOTE}) task. ASOTE also extracts aspect, sentiment and opinion triplets. In the triplet extracted by ASOTE the sentiment is the sentiment of the aspect and opinion pair.  The fourth column in Figure~\ref{fig:asote_aste}  shows the extraction results of the ASOTE task from the corresponding sentences. In addition, we build four datasets for ASOTE based on several popular ABSA benchmarks. 

We propose a Position-aware BERT-based Framework (PBF) to address ASOTE. PBF first extracts aspects from sentences. For each extracted aspect, PBF then extracts associated opinions and predicts the sentiments of the aspect and opinion pairs. PBF obtains triplets by merging the results. Since a sentence may contain multiple aspects associated with different opinions, to extract the corresponding opinions of a given aspect, similar to previous models proposed for the TOWE task~\citep{fan2019target, wu2020latent,pouran-ben-veyseh-etal-2020-introducing,jiang-etal-2021-attention}, PBF generates aspect-specific sentence representations. To accurately generate aspect-specific sentence representations, both the meaning and the position of the aspect are important. Some methods has been proposed to integrate the position information of aspects into non-BERT based models for some ABSA subtasks, such as, ~\citet{gu-etal-2018-position,li-etal-2018-hierarchical} for ATSA, however, how to integrate the position information of aspects into BERT~\citep{devlin2019bert} based modes has not been studied well. PBF generates aspect-specific sentence representations considering both the meaning and the position of the aspect. We explore several methods integrating the position information of aspects into PBF.

Our contributions are summarized as follows:
\begin{itemize}
	\item We introduce a new aspect based sentiment analysis subtask: Aspect-Sentiment-Opinion Triplet Extraction (ASOTE).
	\item We build four datasets for ASOTE and release the datasets for public use as a benchmark.
	\item We propose a Position-aware BERT-based Framework (PBF) to address ASOTE.
	\item Experimental results on the four datasets demonstrate the effectiveness of our method.
\end{itemize}

\section{Related Work}
Aspect-based sentiment analysis (\textbf{ABSA}) \citep{10.1145/1014052.1014073, pontiki-etal-2014-semeval, pontiki-etal-2015-semeval, pontiki-etal-2016-semeval} is a fine-grained sentiment analysis task. ABSA has many subtasks, such as, Aspect Term Extraction (\textbf{ATE}), Opinion Term Exaction (\textbf{OTE}) \footnote{OTE extracts opinions from sentences.}, Aspect Term Sentiment Analysis (\textbf{ATSA}) and Target-oriented Opinion Words Extraction (\textbf{TOWE}) \citep{fan2019target}. Many methods have been proposed for these subtasks. Most methods only solve one subtask, such as~\citet{qiu2011opinion, yin2016unsupervised, li2018aspect, xu2018double, wei2020don,li-etal-2020-conditional} for ATE, \citet{dong2014adaptive, nguyen-shirai-2015-phrasernn,tang-etal-2016-effective, ma2017interactive, xue-li-2018-aspect, jiang-etal-2019-challenge, tang2020dependency, wang2020relational, zhao2020modeling,dai2021does} for ATSA and \citet{fan2019target, wu2020latent,pouran-ben-veyseh-etal-2020-introducing,jiang-etal-2021-attention} for TOWE. Some studies also attempted to solve two or three of these subtasks jointly.  \citet{zhou2019span, li2019unified, phan2020modelling} jointly modeled ATE and ATSA, then generated aspect-sentiment pairs. \citet{wang2016recursive, wang2017coupled, dai2019neural}\footnote{More analysis about the works of \citet{wang2016recursive,wang2017coupled}  can be found in the Appendix A.} jointly modeled ATE and OTE, then output aspect set and opinion set. The extracted
aspects and opinions are not in pairs. \citet{zhao2020spanmlt, chen2020synchronous} jointly modeled ATE and TOWE, then generated aspect-opinion pairs.  \citet{he2019interactive, chen2020relation} jointly modeled ATE, OTE and ATSA, then output aspect-sentiment pairs and opinion  set. However, the extracted
aspects and opinions are also not in pairs, that is, the aspects, sentiments and opinions do not form triplets. The Aspect Sentiment Triplet Extraction (\textbf{ASTE}) task proposed by \citet{Peng2020KnowingWH} extracted aspects, the sentiments of the aspects, and opinions which could form triplets. Some methods have been proposed for ASTE~\citep{xu-etal-2020-position, 10.1007/978-3-030-61609-0_52, 10.1007/978-3-030-60450-9_52, wu-etal-2020-grid,zhang-etal-2020-multi-task,mao2021joint,chen2021bidirectional}  \footnote{More analysis about the work of \citet{zhang-etal-2020-multi-task}  can be found in the Appendix A.}. However, ASTE has the problem mentioned in Section~\ref{sec:introduction}. 

\begin{table*}
	\centering
	\begin{tabular}{|l|l|l|l|l|l|l|l|l|l|l|}
		\hline
		\multicolumn{2}{|l|}{Dataset}  & \#sentence & \#aspects & \#triplets & \#zero\_t & \#one\_t & \#m\_t & \#d\_s1 & \#d\_s2 & \#t\_d \\ \hline
		\multirow{3}{*}{14res} & train & 2429       & 2984      & 2499       & 1662      & 1834     & 304        & 45      & 39      & 181    \\ \cline{2-11} 
		& dev   & 606        & 710       & 561        & 412       & 446      & 54         & 5       & 10      & 24     \\ \cline{2-11} 
		& test  & 800        & 1134      & 1030       & 464       & 720      & 144        & 14      & 9       & 42     \\ \hline
		\multirow{3}{*}{14lap} & train & 2425       & 1927      & 1501       & 1868      & 1128     & 176        & 22      & 26      & 92     \\ \cline{2-11} 
		& dev   & 608        & 437       & 347        & 444       & 268      & 37         & 2       & 2       & 10     \\ \cline{2-11} 
		& test  & 800        & 655       & 563        & 553       & 411      & 69         & 9       & 9       & 40     \\ \hline
		\multirow{3}{*}{15res} & train & 1050       & 950       & 1031       & 471       & 721      & 143        & 22      & 11      & 46     \\ \cline{2-11} 
		& dev   & 263        & 249       & 246        & 134       & 182      & 30         & 4       & 4       & 9      \\ \cline{2-11} 
		& test  & 684        & 542       & 493        & 390       & 385      & 51         & 13      & 5       & 26     \\ \hline
		\multirow{3}{*}{16res} & train & 1595       & 1399      & 1431       & 793       & 1032     & 186        & 35      & 17      & 74     \\ \cline{2-11} 
		& dev   & 400        & 344       & 333        & 209       & 252      & 37         & 4       & 3       & 7      \\ \cline{2-11} 
		& test  & 675        & 612       & 524        & 412       & 395      & 61         & 14      & 6       & 28     \\ \hline
	\end{tabular}
	\caption{\label{ASOTE-datasets} Statistics of our ASOTE datasets. \#zero\_t, \#one\_t and \#m\_t represent the number of aspects without triplet, with one triplet and with multiple triplets, respectively. \#d\_s1 represents the number of aspects that have multiple triplets with different sentiments. \#d\_s2 represents the number of aspects which only have one triplet and whose sentiments are not conflict and are different from the sentiment of the corresponding triplet. \#t\_d represents the number of the triplets whose sentiments are different from the sentiments of the aspects in them.}
\end{table*}

\section{Dataset Construction}
\paragraph{Data Collection} We annotate four datasets (i.e., 14res, 14lap, 15res, 16res) for our propsoed Aspect-Sentiment-Opinion Triplet Extraction (\textbf{ASOTE}) task. First, we construct four Aspect Sentiment Triplet Extraction (\textbf{ASTE}) datasets. Similar to previous studies ~\cite{Peng2020KnowingWH,xu-etal-2020-position}, we obtain four ASTE datasets by aligning the four SemEval Challenge datasets~\citep{pontiki-etal-2015-semeval, pontiki-etal-2016-semeval} and the four Target-oriented Opinion Words Extraction (\textbf{TOWE}) datasets~\citep{fan2019target}. The four SemEval Challenge datasets are restaurant and laptop datasets from SemEval 2014, and restaurant datasets from SemEval 2015 and SemEval 2016. The four SemEval Challenge datasets provide the annotation of aspect terms and the corresponding sentiments, and  the four TOWE datasets were obtained  by annotating the corresponding opinion terms for the annotated aspect terms in the four SemEval Challenge datasets. Compared with the ASTE datasets used in previous  studies~\cite{Peng2020KnowingWH, xu-etal-2020-position},  the ASTE datasets we generate, 1) keep the triplets with conflict sentiments, 2) keep all the sentences in the four SemEval Challenge datasets. That is, the sentences, which don't contain triplets and therefore are not included in the ASTE datasets used in previous  studies~\cite{Peng2020KnowingWH, xu-etal-2020-position}, are included in the ASTE datasets we generate. We think datasets including these sentences can better evaluate the performance of ASOTE methods, since ASOTE methods can encounter this kind of sentences in real-world scenarios. 

\begin{figure}
	\centering
	\includegraphics[scale=0.17]{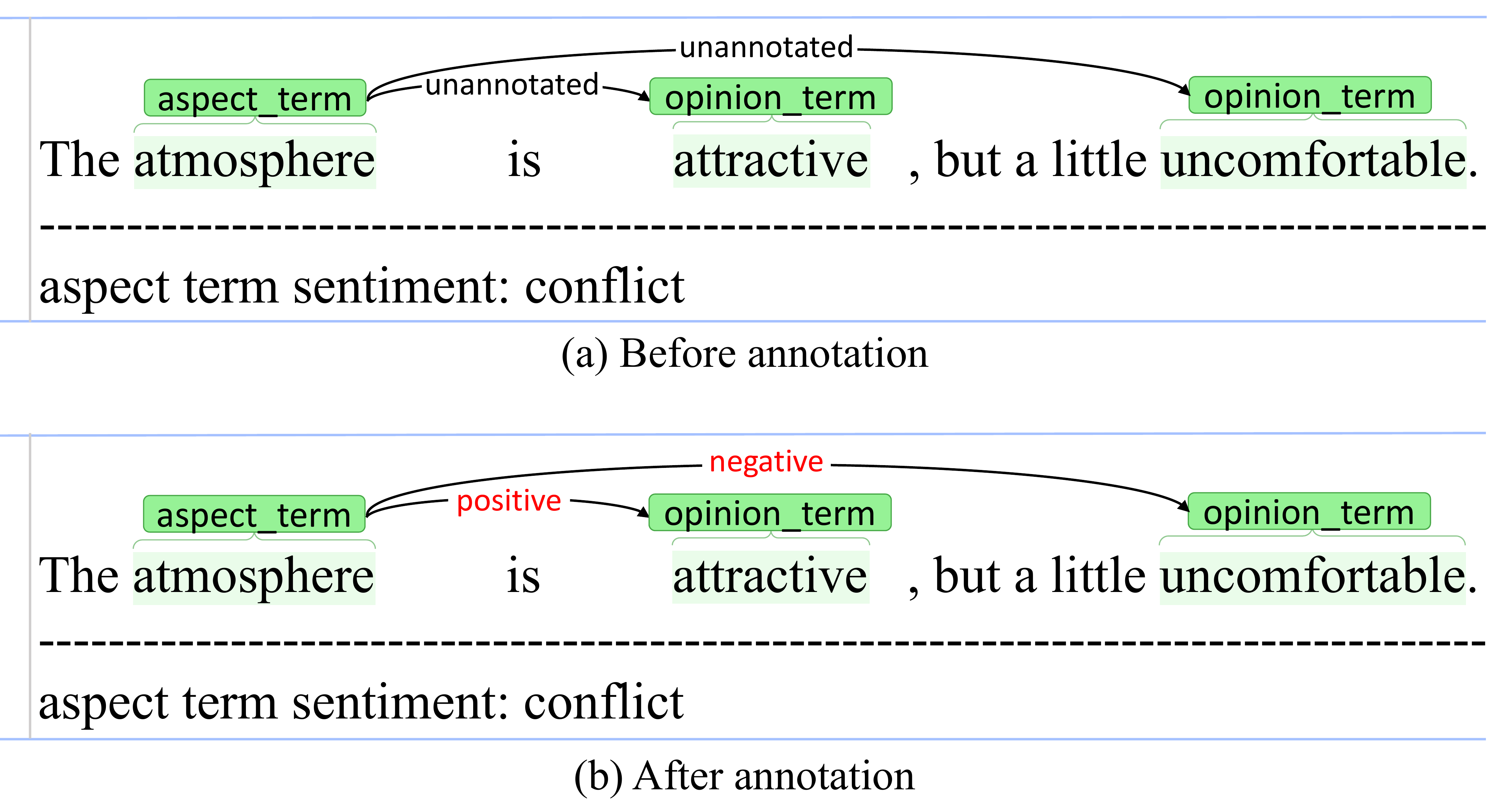}
	\caption{An example of annotating the sentiments of the aspect and opinion pairs on the ASTE triplets for the ASOTE task.}
	\label{fig:annotation}
\end{figure}

\paragraph{Data Annotation} We invited a researcher who works on natural language processing (NLP) and an undergraduate to annotate the sentiments of the aspect-opinion pairs in the triplets of the four ASTE datasets. The annotation tool we used is brat~\citep{stenetorp-etal-2012-brat}. Each time, we only provided triplets of one aspect term to the annotators. For each aspect term, not only the aspect term and its corresponding opinion terms but also the sentiment of the aspect term were provided to the annotators. Figure~\ref{fig:annotation} (a) shows an example of what we provided to the annotators and Figure~\ref{fig:annotation} (b) shows the results of annotation. When annotating the sentiment of an aspect-opinion pair, the annotators need to consider both the opinion itself and the context of the opinion. For example, given the sentence, ``The decor is night tho...but they REALLY need to clean that vent in the ceiling...its quite un-appetizing, and kills your effort to make this place look sleek and modern.''\footnote{The triplets extracted by ASOTE from this sentence, i.e. (``place'', negative, ``sleek'') and (``place'', negative, ``modern''), are also confusing, since the sentiment shifter expression is complicated and therefore is not annotated as part of the opinions. One simple solution to this problem is to add a reversing word (e.g.,``not'') to this kind of opinions (e.g., ``not sleek'' and ``not modern'') when we annotate opinions, which is left for future exploration.} and one aspect-opinion pair, (``place'', ``sleek''), the sentiment should be negative, even though the sentiment of ``sleek'' is positive. The kappa statistic~\citep{doi:10.1177/001316446002000104} between the annotations of the two annotators is 0.85. The conflicts have be checked by another researcher who works on  NLP.

\paragraph{Dataset Analysis} The statistics of the four ASOTE datasets are summarized in Table~\ref{ASOTE-datasets}. Since \#diff\_s2 is always greater than 0,  the annotators have to annotate  the sentiments of the triplets in which the aspect only have one triplet and the sentiment of the aspect is not conflict. That is, we can not treat the sentiment of the aspect in these triplets as the sentiment of these triplets. For example, for the third sentence in Figure~\ref{fig:asote_aste}, the aspect ``Food'' has negative sentiment, while the correct sentiment of its only one triplet, (``Food'', neutral, ``average''), is neutral.

\section{Method}
In this section, we describe our Position-aware BERT-based Framework (\textbf{PBF}) for Aspect-Sentiment-Opinion Triplet Extraction (\textbf{ASOTE}). 

\subsection{Task Definition}
Given a sentence $S=\{w_0,...,w_i,...,w_{n-1}\}$ containing $n$ words, ASOTE aims to extract a set of triplets: $T=\{(a, s, o)_t\}_{t=0}^{|T|-1}$, where $a$ is an aspect, $o$ is an opinion, $s$ is the sentiment of the aspect-opinion pair $(a, o)$, and $|T|$ is the number of triplets in the sentence. When a sentence does not contain triplets,  $|T| = 0$.

\subsection{PBF}
\begin{figure}
	\centering
	\includegraphics[scale=0.152]{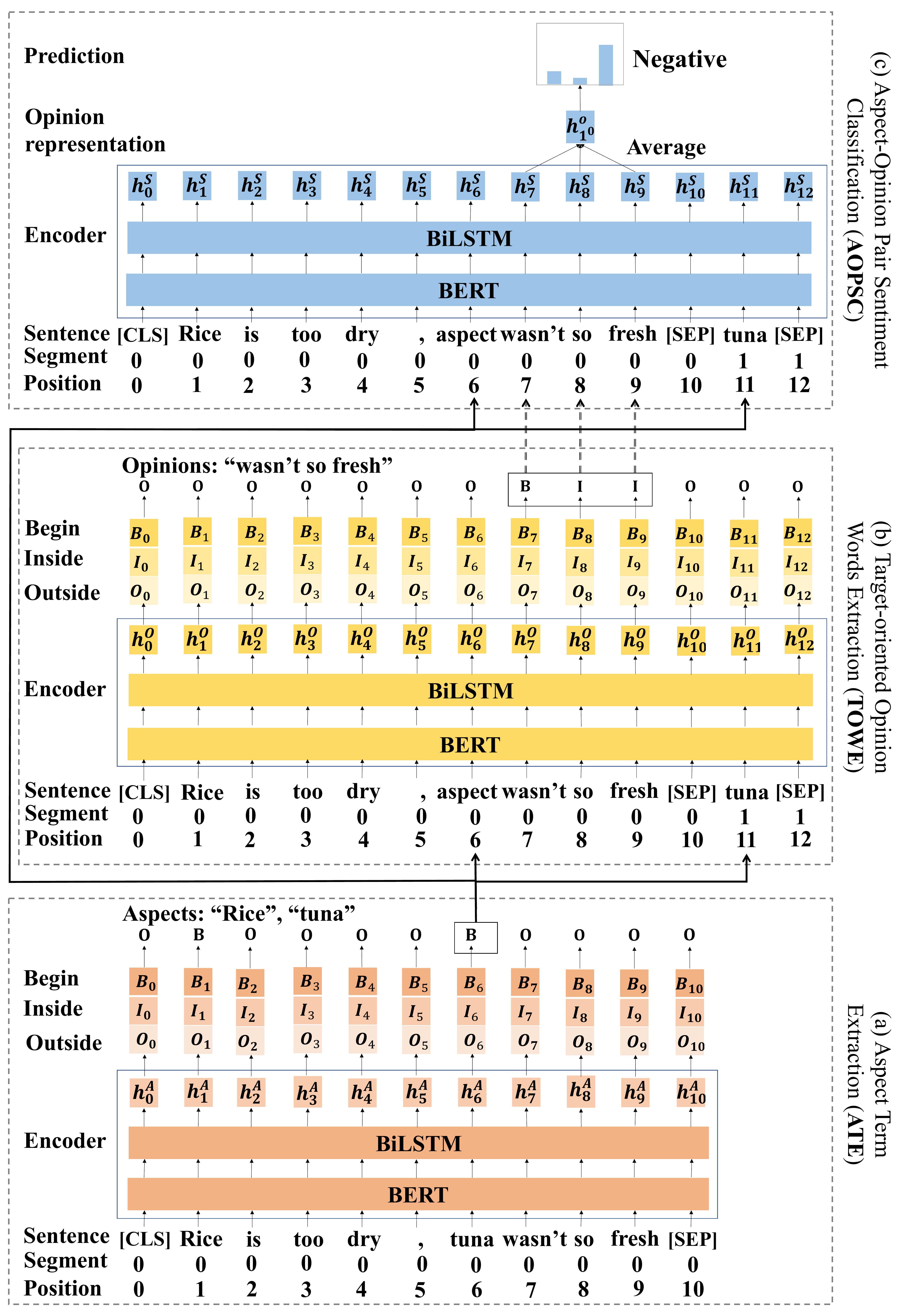}
	\caption{Our proposed Position-aware BERT-based Framework (\textbf{PBF}).}
	\label{fig:PBF}
\end{figure}

Figure~\ref{fig:PBF} shows the overview of PBF. PBF contains three models. Given a sentence $S=\{w_0,...,w_i,...,w_{n-1}\}$, the Aspect Term Extraction (\textbf{ATE}) model first extracts a set of aspects $A=\{a_0,...,a_j,...,a_{m-1}\}$. For each extracted aspect, $a_j$, the Target-oriented Opinion Words Extraction (\textbf{TOWE}) model then extracts its opinions $O=\{o_j^0,...,o_j^k,..., o_j^{{l_j}-1}\}$, where $l_j$ is the number of opinions with respect to the $j$-th aspect and $l_j \ge 0$. Finally, for each extracted aspect-opinion pair $(a_j, o_j^k)$, the Aspect-Opinion Pair Sentiment Classification (\textbf{AOPSC}) model predicts its sentiment  $s_j^k \in P = \{positive, neutral, negative\}$. PBF obtains the triplets by merging the results of the three models: $T=\{(a_0, s_0^0, o_0^0),..., (a_{m-1}, s_{m-1}^{l_{m-1}}, o_{m-1}^{l_{m-1}})\}$. In PBF, all three models use BiLSTM~\citep{graves2013speech} with BERT~\citep{devlin2019bert} as sentence encoder.

\subsection{Input}
The input of PBF is a sentence $S=\{w_0,...,w_i,...,w_{n-1}\}$ consisting of $n$ words.

Given the sentence, to obtain the inputs of the ATE model (Figure~\ref{fig:PBF} (a)), we first convert the sentence $S$ to a new sentence, $S_B=\{w_0,...,w_i,...,w_q\}$, where $w_0$ is "[CLS]" and $w_q$ is "[SEP]". We then generate segment indices $I_{seg}=\{0, ...,0\}$ and position indices $I_{pos}=\{0, ..., q\}$ for the new sentence. 

Since a sentence may contain multiple aspects associated with different opinions, to extract the associated opinions of a particular aspect, the TOWE model generates aspect-specific sentence representations for the aspect. It's intuitive that both the meaning and the position of the aspect are important for producing aspect-specific sentence representations. In other words, we need to tell the TOWE model what the aspect is and where the aspect is in the sentence. Given the sentence $S$ and an aspect $a_j$ in the sentence, we first replace the words of the aspect with the word ``aspect'', which tells the TOWE model where the aspect is in the sentence. We then append the words of the aspect to the end of the sentence, which tells the model what the aspect is. Finally, we obtain a new sentence $S^A_B=\{w_0,...,w_i,...,w_q\}$. We also generate segment indices $I^A_{seg}=\{0, ...,1\}$ and position indices $I^A_{pos}=\{0, ..., q\}$ for the new sentence. The encoder of the TOWE model (Figure~\ref{fig:PBF} (b)) takes $S^A_B$, $I^A_{seg}$ and $I^A_{pos}$ as inputs, and can generate aspect-specific sentence representations. 

To predict the sentiment of an aspect-opinion pair, the AOPSC model (Figure~\ref{fig:PBF} (c)) also generate aspect-specific sentence representations for the aspect. The inputs of the AOPSC model are the same as the TOWE model.

\subsection{ATE}
We formulate ATE as a sequence labeling problem. The encoder takes $S_B$, $I_{seg}$ and $I_{pos}$ as inputs, and outputs the corresponding sentence representation, $H^A=\{h^A_0,...,h^A_i,...,h^A_q\}$. ATE model uses $h^A_i$ to predict the tag $y^A_i \in \{B, I, O\}$ (B: Begin, I: Inside, O: Outside) of the word $w_i$. It can be regarded as a three-class classification problem at
each position of $S_B$. We use a linear layer and a softmax layer to compute prediction probability $\hat{y}^A_i$:
\begin{equation}
	\hat{y}^A_i=softmax(W^A_1h^A_i + b^A_1)
\end{equation}
where $W^A_1$ and $b^A_1$ are learnable parameters.

The cross-entropy loss of ATE task can be defined as follows:
\begin{equation}
	L_{ATE}=-\sum_{i=0}^{q}\sum_{t \in \{B, I, O\}}\mathbb{I}(y^A_i=t)log(\hat{y}^A_{i_t})
\end{equation}
where $y^A_i$ denotes the ground truth label. $\mathbb{I}$ is an indicator function.  If $y^A_i==t$, $\mathbb{I}$ = 1, otherwise $\mathbb{I}$ = 0. We minimize $L_{ATE}$ to optimize the ATE model.

Finally, ATE model decodes the tag sequence of the sentence and outputs a set of aspects $A=\{a_0,...,a_j,...,a_{m-1}\}$.

\subsection{TOWE}
We aslo formulate TOWE as a sequence labeling problem. The TOWE model has the same architecture as the ATE model, but they do not share the parameters. The TOWE model takes $S^A_{B}$, $I^A_{seg}$ and $I^A_{pos}$ as inputs and outputs the opinions  $O=\{o_j^0,...,o_j^k,..., o_j^{{l_j}-1}\}$ of the aspect $a_j$.

\subsection{AOPSC}
Given an aspect $a_j$ and its opinions  $\{o_j^0,...,o_j^k,..., o_j^{{l_j}-1}\}$, the AOPSC model predicts the sentiments $\{s_j^0, ..., s_j^k, ..., s_j^{{l_j}-1}\}$ of all aspect-opinion pairs, $\{(a_j, o_j^0),...,(a_j, o_j^k),..., (a_j, o_j^{{l_j}-1})\}$, at once. The encoder of the AOPSC model takes the new sentence $S^A_{B}$, the segment indices $I^A_{seg}$ and the position indices $I^A_{pos}$ as inputs and outputs the aspect-specific sentence representation, $H^S=\{h^S_0,...,h^S_{q}\}$. We then obtain the representation of an opinion by averaging the hidden representations of the words in the opinion. The representation $h^o_{j^k}$ of opinion $o_j^k$ is used to make sentiment prediction $\hat{y}^o_{j^k}$ of opinion $o_j^k$:
\begin{equation}
	\hat{y}^o_{j^k}=softmax(W^S_1h^o_{j^k} + b^S_1)
\end{equation}
where $W^S_1$ and $b^S_1$ are learnable parameters.

The loss of the AOPSC task is the sum of all opinions' cross entropy of the aspect:
\begin{equation}
	L_{AOPSC}=-\sum_{k=0}^{{l_j}-1}\sum_{t \in P}\mathbb{I}(y^o_{j^k}=t)log\hat{y}^o_{j^k_t}
\end{equation}
where $y^o_{j^k}$ denotes the ground truth label. We minimize $L_{AOPSC}$ to optimize the AOPSC model. 

\section{Experiments}
\subsection{Datasets and Metrics}
We evaluate our method on two types of datasets:

\textbf{TOWE-data} \citep{fan2019target} is used to compare our method with previous methods proposed for the Target-oriented Opinion Words Extraction (\textbf{TOWE}) task on the TOWE task. TOWE-data only includes the sentences that contain pairs of aspect and opinion and the aspect associated with at least one opinion. Following previous works \citep{fan2019target, wu2020latent},  We randomly select 20\% of training set as development set for tuning hyper-parameters and early stopping. 

\textbf{ASOTE-data} is the data we build for our Aspect-Sentiment-Opinion Triplet Extraction (\textbf{ASOTE}) task and is used to compare the methods on the ASOTE task. ASOTE-data can also be used to evaluate the TOWE models on the TOWE task.  Compared with TOWE-data, ASOTE-data additionally includes the sentences that do not contain aspect-opinion pairs and includes the aspects without opinions. Since methods can encounter these kind of examples in real-world scenarios, ASOTE-data is more appropriate to evaluate methods on the TOWE task. 

We use precision (P), recall (R), and F1-score (F1) as the evaluation metrics. For the ASOTE task, an extracted triplet is regarded as correct only if predicted aspect spans, sentiment, opinion spans and ground truth aspect spans, sentiment, opinion spans are exactly matched.

\subsection{Our Methods}

\begin{figure}
	\centering
	\includegraphics[scale=0.152]{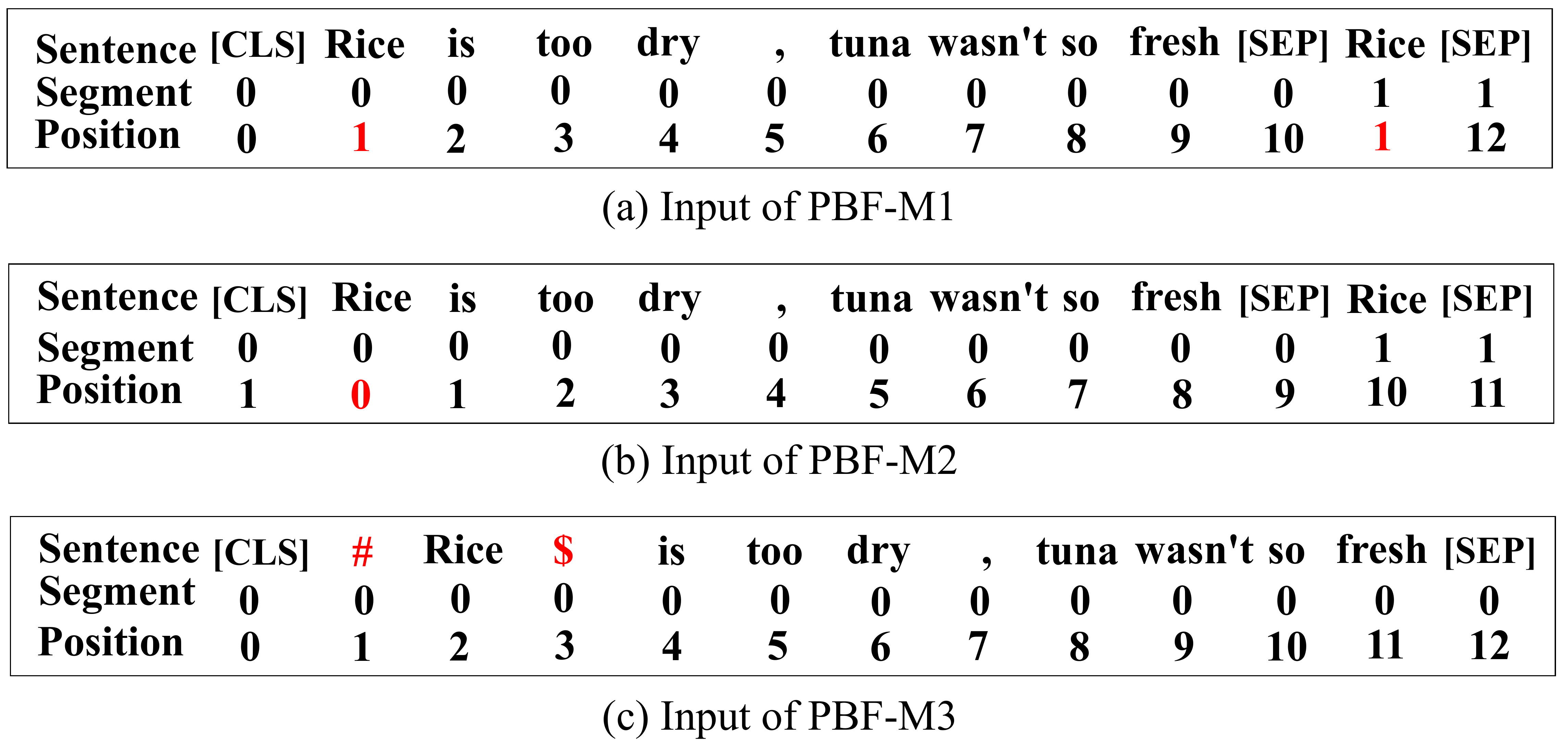}
	\caption{The inputs of PBF-M1, PBF-M2 and PBF-M2, given the sentence ``Rice is too dry, tuna was n't so fresh'' and the aspect ``Rice''.}
	\label{fig:inputs}
\end{figure}

\begin{table*}
	\centering
	\begin{tabular}{|l|lll|lll|lll|lll|}
		\hline
		& \multicolumn{3}{c|}{14res}                                              & \multicolumn{3}{c|}{14lap}                                              & \multicolumn{3}{c|}{15res}                                              & \multicolumn{3}{c|}{16res}                                              \\ \hline
		Method          & \multicolumn{1}{c}{P} & \multicolumn{1}{c}{R} & \multicolumn{1}{c|}{F1} & \multicolumn{1}{c}{P} & \multicolumn{1}{c}{R} & \multicolumn{1}{c|}{F1} & \multicolumn{1}{c}{P} & \multicolumn{1}{c}{R} & \multicolumn{1}{c|}{F1} & \multicolumn{1}{c}{P} & \multicolumn{1}{c}{R} & \multicolumn{1}{c|}{F1} \\
		OTE-MTL         & 63.8                  & 52.1                  & 57.3                    & 51.3                  & 36.8                  & 42.7                    & 56.3                  & 44.0                  & 49.3                    & 58.3                  & 52.4                  & 55.0                    \\
		JET$^t$         & 66.0                  & 48.4                  & 55.8                    & 41.0                  & 36.5                  & 38.6                    & 45.3                  & 47.8                  & 46.5                    & 58.1                  & 46.9                  & 51.9                    \\
		JET$^o$         & 61.5                  & 53.9                  & 57.5                    & 48.8                  & 36.7                  & 41.9                    & 57.5                  & 47.2                  & 51.8                    & 61.0                  & 56.8                  & 58.8                    \\
		GTS-CNN         & 66.4                  & 58.5                  & 62.2                    & 55.3                  & 37.4                  & 44.6                    & 56.3                  & 48.1                  & 51.8                    & 61.4                  & 60.0                  & 60.5                    \\
		GTS-BiLSTM      & 71.1                  & 54.5                  & 61.5                    & 58.0                  & 33.9                  & 42.8                    & 67.3                  & 42.9                  & 52.4                    & 64.6                  & 55.8                  & 59.8                    \\
		JET$^t_{+bert}$ & 65.1                  & 51.7                  & 57.6                    & 50.2                  & 41.7                  & 45.5                    & 50.7                  & 48.2                  & 49.4                    & 55.0                  & 52.1                  & 53.5                    \\
		JET$^o_{+bert}$ & 66.0                  & 54.5                  & 59.7                    & 49.7                  & 42.8                  & 46.0                    & 53.8                  & 52.9                  & 53.3                    & 58.3                  & 60.3                  & 59.2                    \\
		GTS-BERT        & 67.5                  & 67.2                  & 67.3                    & 59.4                  & 48.6                  & 53.5                    & 61.8                  & 52.0                  & 56.4                    & 62.0                  & 67.1                  & 64.4                    \\ \hline
		PBF             & 69.3                  & 69.0                  & {\ul \textbf{69.2}}     & 56.6                  & 55.1                  & \textbf{55.8}           & 55.8                  & 61.5                  & \textbf{58.5}           & 61.2                  & 72.7                  & {\ul \textbf{66.5}}     \\
		PBF -w/o A      & 67.3                  & 69.3                  & 68.3                    & 55.9                  & 55.7                  & 55.8                    & 56.4                  & 61.6                  & {\ul 58.8}              & 60.7                  & 71.3                  & 65.5                    \\
		PBF -w/o P      & 68.6                  & 69.7                  & 69.1                    & 56.6                  & 54.8                  & 55.7                    & 56.2                  & 60.4                  & 58.2                    & 59.6                  & 71.8                  & 65.1                    \\
		PBF -w/o AP     & 44.4                  & 51.9                  & 47.4                    & 45.1                  & 48.8                  & 46.7                    & 41.7                  & 50.9                  & 45.7                    & 46.1                  & 59.8                  & 52.0                    \\
		PBF-M1          & 66.6                  & 69.7                  & 68.1                    & 58.8                  & 54.1                  & {\ul 56.3}              & 57.8                  & 59.4                  & 58.4                    & 59.3                  & 72.1                  & 65.0                    \\
		PBF-M2          & 63.0                  & 63.6                  & 63.3                    & 51.8                  & 47.3                  & 49.4                    & 50.2                  & 56.2                  & 53.0                    & 56.6                  & 65.8                  & 60.8                    \\
		PBF-M3          & 66.8                  & 69.2                  & 68.0                    & 56.8                  & 53.3                  & 54.9                    & 54.2                  & 61.7                  & 57.7                    & 60.4                  & 71.1                  & 65.2                    \\ \hline
	\end{tabular}
	\caption{\label{table:ASOTE} Results of ASOTE task. The bold F1 scores are the best scores among PBF and the baselines. The underlined F1 scores are the best scores among PBF and its variants.}
\end{table*}

\begin{table}
	\centering
	\begin{tabular}{|l|l|l|l|l|}
		\hline
		Method   & 14res       & 14lap         & 15res         & 16res         \\ \hline
		GTS-BERT & 71.7        & 60.2          & 61.5          & 68.1          \\ \hline
		PBF      & \textbf{74.0} & \textbf{63.8} & \textbf{63.9} & \textbf{70.8} \\ \hline
	\end{tabular}
	\caption{\label{table:AOPE} Results of the OPE task in terms of F1.}
\end{table}

We provide the comparisons of several variants of our Position-aware BERT-based Framework (\textbf{PBF}).
The difference between these variants is the way they generate the new sentence $S^A_B$, the segment indices $I^A_{seg}$ and the position indices $I^A_{pos}$.

\textbf{PBF -w/o A} doesn't append the words of the aspect to the end of the original sentence. In other words, this variant doesn't know what the aspect is.

\textbf{PBF -w/o P} does not replace the words of the aspect with the word ``aspect'', namely that this variant does not know where the aspect is. This model has been used on some aspect-based sentiment analysis subtasks to generate aspect-specific sentence representations~\citep{jiang-etal-2019-challenge,li-etal-2020-multi-instance}.

\textbf{PBF -w/o AP} neither appends the words of the aspect to the end of the original sentence, nor replaces the words of the aspect with the word ``aspect''.

\textbf{PBF-M1} does not replace the words of the aspect with the word ``aspect''. In order to tell the model the position of the aspect, the words of the aspect in the original sentence and the words of the aspect appended to the original sentence have the same position indices. This method has been used on relation classification~\citep{zhong2020frustratingly}.

\textbf{PBF-M2} does not replace the words of the aspect with the word ``aspect''. In order to tell the model the position of the aspect, the position indices of the words of the aspect in the original sentence are marked as 0, and the position indices of other words are the relative distance to the aspect. This method has been used on the aspect-term sentiment analysis task~\citep{gu-etal-2018-position}.

\textbf{PBF-M3} modifies the original sentence $S$ by inserting the special token \# at the beginning of the aspect and the special token \$ at the end of the aspect. Special tokens were first used by \citet{wu2019enriching} to incorporate target entities information into BERT on the relation classification task. 

Figure~\ref{fig:inputs} shows inputs examples for  PBF-M1, PBF-M2 and PBF-M3.

\subsection{Implementation Details}
We implement our models in PyTorch \citep{paszke2017automatic}. We use the uncased basic pre-trained BERT. The BERT is fine-tuned during training. The batch size is set to 32 for all models. All models are optimized by the Adam optimizer \citep{kingma2014adam}. The learning rates is 0.00002. we apply a dropout of $p= 0.5$ after the BERT and BiLSTM layers. We apply early stopping in training and the patience is 10. We run all models for 5 times and report the average results on the test datasets. For the baseline models of the ASOTE task, we first convert our datasets to the datasets which has the same formats as the inputs of the baseline models, then run the code that the authors released on the converted datasets. 

\subsection{Exp-I: ASOTE}
\subsubsection{Comparison Methods}
On the ASOTE task,we compare our methods with several methods proposed for the Aspect Sentiment Triplet Extraction (\textbf{ASTE}) task. These methods also extract aspect, sentiment, opinion triplets from sentences. These methods include MTL from \citet{zhang-etal-2020-multi-task}\footnote{https://github.com/l294265421/OTE-MTL-ASOTE}, JET$^t$, JET$^o$, JET$^t_{+bert}$ and JET$^o_{+bert}$  where  $M=6$ from \citet{xu-etal-2020-position}\footnote{https://github.com/l294265421/Position-Aware-Tagging-for-ASOTE}, GTS-CNN, GTS-BiLSTM and GTS-BERT from \citet{wu-etal-2020-grid}\footnote{https://github.com/l294265421/GTS-ASOTE}. All these baselines are joint models, which are jointly trained to extract the three elements of ASOTE triplets.

\begin{figure*}
	\centering
	\includegraphics[scale=0.165]{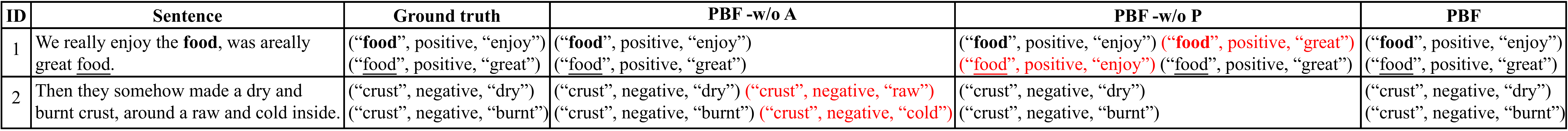}
	\caption{Case study. Red triplets are incorrect predictions.}
	\label{fig:case_study}
\end{figure*}

\subsubsection{Results}
\label{asote_result}
The results of the ASOTE task are
shown in Table~\ref{table:ASOTE}. We have several observations from Table~\ref{table:ASOTE}. First, MTL outperforms JET$^t$ on all datasets, because JET$^t$ can extract at most one triplet for an aspect. Although JET$^o$  can extract at most one triplet for an opinion, JET$^o$ outperforms JET$^t$ on all datasets and surpasses MTL on 3 of 4 datasets, because the opinions belonging to multiple triplets are less than the aspects belonging to multiple triplets\footnote{More statistics about our ASOTE datasets can be found in the Appendix B.}. Second, GTS-CNN and GTS-BiLSTM outperform  both JET$^t$ and JET$^o$ on all datasets, and GTS-BERT also achieves better performance than JET$^t_{+bert}$ and JET$^o_{+bert}$. GTS-BERT is the best baseline model. Third, our proposed PBF surpasses GTS-BERT on all datasets. Since the Aspect Term Extraction (\textbf{ATE}) model and the Aspect-Opinion Pair Sentiment Classification (\textbf{AOPSC}) model in PBF are vanilla\footnote{The results of PBF and its variants on AOPSC can be found in the Appendix C. Compared with PBF -w/o AP, PBF and the other variants can't obtain better performance consistently on all datasets.},  compared with previous models, the advantages of PBF are from the TOWE model. However,  GTS-BERT can not be applied to the TOWE task directly, so we compare PBF with GTS-BERT on the aspect-Opinion Pair Extraction (\textbf{OPE})~\citep{wu-etal-2020-grid} task. The results of OPE are shown in Table~\ref{table:AOPE}, which shows that PBF also outperforms GTS-BERT on all datasets. Fourth, PBF outperforms PBF -w/o P on all datasets, indicating that integrating position information of aspects can boost the model performance. Fifth, PBF -w/o A surpasses PBF -w/o P on the 14lap, 15res and 16res datasets, which indicates that the position information of aspects is more effective than the meanings of aspects on these datasets. Sixth, PBF outperforms PBF-M1 and PBF-M2, which shows that our method that tells the model where the aspect is is more effective. Although the mehtod used by PBF-M2 to integrate the position information of aspects into it has been successfully applied to non-BERT based models, it is not effective enough for BERT-based models. Seventh, PBF outperforms PBF-M3, indicting our method is more effective than the method that integrating the position information and meaning of an aspect into a model by inserting special aspect markers for the aspect. The possible reason is that the additional
special tokens may destroy the syntax knowledge
learned by BERT. Last but not least, PBF -w/o AP obtains the worst performance among all variants, which further demonstrates that both the position and the meaning of an aspect are important.

\subsubsection{Case  Study}
To further understand the effect of the position and the meaning of an aspect, we perform a case study on two sentences, as displayed in Figure~\ref{fig:case_study}. In the first sentence, the bold ``food'' and underlined ``food'' are different aspects. The positions of the aspects help PBF and PBF -w/o A to extract different opinions for aspects with the same meaning. In the second sentence, with the help of the meaning of the aspect ``crust'', PBF and PBF -w/o P do not extract ``raw'' and ``cold'' as the opinions of ``crust''.

\begin{table}
	\centering
	\begin{tabular}{|l|l|l|l|l|}
		\hline
		Method      & 14res         & 14lap         & 15res         & 16res         \\ \hline
		PBF         & \textbf{81.5} & 74.0            & 77.9          & \textbf{82.1} \\ \hline
		PBF -w/o A  & 80.7          & \textbf{74.1} & \textbf{78.6} & 81.6          \\ \hline
		PBF -w/o P  & 80.9          & 74.0            & 77.3          & 81.0            \\ \hline
		PBF -w/o AP & 56.1          & 61.9          & 60.5          & 64.8          \\ \hline
		PBF-M1      & 80.1          & 73.0            & 77.4          & 80.5          \\ \hline
		PBF-M2      & 75.1          & 66.1          & 72.9          & 76.5          \\ \hline
		PBF-M3      & 80.3          & 73.6          & 77.5          & 80.3          \\ \hline
	\end{tabular}
	\caption{\label{table:TOWE-ASOTE-data} Results of the TOWE task in terms of F1 on the ASOTE-data. }
\end{table}

\subsection{Exp-II: TOWE}
\subsubsection{Comparison Methods}
On the TOWE task, we compare our methods with (1) three non-BERT models: \textbf{IOG}~\citep{fan2019target}, \textbf{LOTN}~\citep{wu2020latent}, \textbf{ARGCN}~\citep{jiang-etal-2021-attention} (2) two BERT-based models: \textbf{ARGCN$_{+bert}$}~\citep{jiang-etal-2021-attention} and \textbf{ONG}~\citep{pouran-ben-veyseh-etal-2020-introducing}.

\begin{table}
	\centering
	\begin{tabular}{|l|l|l|l|l|}
		\hline
		Method          & 14res         & 14lap         & 15res         & 16res         \\ \hline
		IOG             & 80.0         & 71.3         & 73.2         & 81.6         \\ \hline
		LOTN            & 82.2         & 72.0         & 73.2         & 83.6         \\ \hline
		ARGCN           & 84.6         & 75.3         & 76.7         & 85.1         \\ \hline
		ARGCN$_{+bert}$ & 85.4         & 76.3         & 78.2         & 86.6         \\ \hline
		ONG             & 82.3         & 75.7         & 78.8         & 86.0         \\ \hline
		PBF             & 85.9          & \textbf{81.5} & \textbf{80.8} & \textbf{89.2} \\ \hline
		PBF -w/o A      & 86.1          & 81.2          & 80.4          & 87.9          \\ \hline
		PBF -w/o P      & \textbf{86.3} & 80.3          & 79.8          & 88.8          \\ \hline
		PBF -w/o AP     & 61.6          & 67.9          & 59.0            & 69.3          \\ \hline
	\end{tabular}
	\caption{\label{table:TOWE-TOWE-data} Results of the TOWE task in terms of F1 on the TOWE-data. The results of the baselines cite form the original papers.}
\end{table}

\subsubsection{Results}
The results on ASOTE-data are shown in Table~\ref{table:TOWE-ASOTE-data} and the results on TOWE-data are shown in Table~\ref{table:TOWE-TOWE-data}. We
draw the following conclusions from the results.  First, PBF outperforms all baselines proposed for TOWE on the TOWE-data, indicating the effectiveness of our method. Second, PBF -w/o P also surpasses all baselines on the TOWE-data. To the best of our knowledge, no previous study evaluates the performance of this method on TOWE. Third, regarding PBF and its variants, we can obtain conclusions from Table~\ref{table:TOWE-ASOTE-data} similar to the conclusions obtained from Table~\ref{table:ASOTE}, because the differences of these models' performance on ASOTE are mainly brought by the differences of their performacnes on TOWE. Fourth, since the
methods (i.e., PBF, PBF -w/o A, PBF -w/o P and PBF -w/o AP) obtain better performance on TOWE-data
than on ASOTE-data, ASOTE-data is a more
challenging dataset for TOWE.

\section{Conclusion}
In this paper, we introduce the Aspect-Sentiment-Opinion
Triplet Extraction (ASOTE) task. ASOTE is more fine-grained
than Aspect Sentiment Triplet Extraction
(ASTE). The sentiment of an triplet extracted by
ASOTE is the sentiment of the aspect-opinion pair
in the triplet. We manually annotate four datasets
for ASOTE. Moreover, We propose a Position-aware BERT-based Framework (PBF) to address ASOTE. Although PBF is a pipeline method, it obtains better performance than several joint models, which
demonstrates the effectiveness of our method.

 Since Aspect Term Extraction (ATE) and Target-oriented Opinion Words Extraction (TOWE) are highly correlated with each other, and TOWE and Aspect-Opinion Pair Sentiment Classification (AOPSC) are also highly correlated with each other, we can improve PBF by turning it into a joint model which jointly trains the ATE model, the TOWE model and the AOPSC model. However, it is not easy to jointly train the ATE model and the TOWE model, since we need to use the aspects that the ATE model extracts to modify the sentences that the TOWE model takes as input. In the future, we will explore how to jointly train the ATE model and the TOWE model.

\bibliography{anthology,custom}

\begin{thebibliography}{55}
\expandafter\ifx\csname natexlab\endcsname\relax\def\natexlab#1{#1}\fi

\bibitem[{Chen et~al.(2020{\natexlab{a}})Chen, Chen, and
  Liu}]{10.1007/978-3-030-60450-9_52}
Peng Chen, Shaowei Chen, and Jie Liu. 2020{\natexlab{a}}.
\newblock Hierarchical sequence labeling model for aspect sentiment triplet
  extraction.
\newblock In \emph{Natural Language Processing and Chinese Computing}, pages
  654--666, Cham. Springer International Publishing.

\bibitem[{Chen et~al.(2020{\natexlab{b}})Chen, Liu, Wang, Zhang, and
  Chi}]{chen2020synchronous}
Shaowei Chen, Jie Liu, Yu~Wang, Wenzheng Zhang, and Ziming Chi.
  2020{\natexlab{b}}.
\newblock Synchronous double-channel recurrent network for aspect-opinion pair
  extraction.
\newblock In \emph{Proceedings of the 58th Annual Meeting of the Association
  for Computational Linguistics}, pages 6515--6524.

\bibitem[{Chen et~al.(2021)Chen, Wang, Liu, and Wang}]{chen2021bidirectional}
Shaowei Chen, Yu~Wang, Jie Liu, and Yuelin Wang. 2021.
\newblock Bidirectional machine reading comprehension for aspect sentiment
  triplet extraction.
\newblock \emph{arXiv preprint arXiv:2103.07665}.

\bibitem[{Chen and Qian(2020)}]{chen2020relation}
Zhuang Chen and Tieyun Qian. 2020.
\newblock Relation-aware collaborative learning for unified aspect-based
  sentiment analysis.
\newblock In \emph{Proceedings of the 58th Annual Meeting of the Association
  for Computational Linguistics}, pages 3685--3694.

\bibitem[{Cohen(1960)}]{doi:10.1177/001316446002000104}
Jacob Cohen. 1960.
\newblock \href {https://doi.org/10.1177/001316446002000104} {A coefficient of
  agreement for nominal scales}.
\newblock \emph{Educational and Psychological Measurement}, 20(1):37--46.

\bibitem[{Dai and Song(2019)}]{dai2019neural}
Hongliang Dai and Yangqiu Song. 2019.
\newblock Neural aspect and opinion term extraction with mined rules as weak
  supervision.
\newblock In \emph{Proceedings of the 57th Annual Meeting of the Association
  for Computational Linguistics}, pages 5268--5277.

\bibitem[{Dai et~al.(2021)Dai, Yan, Sun, Liu, and Qiu}]{dai2021does}
Junqi Dai, Hang Yan, Tianxiang Sun, Pengfei Liu, and Xipeng Qiu. 2021.
\newblock Does syntax matter? a strong baseline for aspect-based sentiment
  analysis with roberta.
\newblock \emph{arXiv preprint arXiv:2104.04986}.

\bibitem[{Devlin et~al.(2019)Devlin, Chang, Lee, and
  Toutanova}]{devlin2019bert}
Jacob Devlin, Ming-Wei Chang, Kenton Lee, and Kristina Toutanova. 2019.
\newblock \href {https://doi.org/https://doi.org/10.18653/v1/N19-1423} {Bert:
  Pre-training of deep bidirectional transformers for language understanding}.
\newblock In \emph{Proceedings of the 2019 Conference of the North American
  Chapter of the Association for Computational Linguistics: Human Language
  Technologies, Volume 1 (Long and Short Papers)}, pages 4171--4186.

\bibitem[{Dong et~al.(2014)Dong, Wei, Tan, Tang, Zhou, and
  Xu}]{dong2014adaptive}
Li~Dong, Furu Wei, Chuanqi Tan, Duyu Tang, Ming Zhou, and Ke~Xu. 2014.
\newblock Adaptive recursive neural network for target-dependent twitter
  sentiment classification.
\newblock In \emph{Proceedings of the 52nd annual meeting of the association
  for computational linguistics (volume 2: Short papers)}, pages 49--54.

\bibitem[{Fan et~al.(2019)Fan, Wu, Dai, Huang, and Chen}]{fan2019target}
Zhifang Fan, Zhen Wu, Xinyu Dai, Shujian Huang, and Jiajun Chen. 2019.
\newblock Target-oriented opinion words extraction with target-fused neural
  sequence labeling.
\newblock In \emph{Proceedings of the 2019 Conference of the North American
  Chapter of the Association for Computational Linguistics: Human Language
  Technologies, Volume 1 (Long and Short Papers)}, pages 2509--2518.

\bibitem[{Graves et~al.(2013)Graves, Mohamed, and Hinton}]{graves2013speech}
Alex Graves, Abdel-rahman Mohamed, and Geoffrey Hinton. 2013.
\newblock \href {https://doi.org/https://doi.org/10.1109/ICASSP.2013.6638947}
  {Speech recognition with deep recurrent neural networks}.
\newblock In \emph{2013 IEEE international conference on acoustics, speech and
  signal processing}, pages 6645--6649. IEEE.

\bibitem[{Gu et~al.(2018)Gu, Zhang, Hou, and Song}]{gu-etal-2018-position}
Shuqin Gu, Lipeng Zhang, Yuexian Hou, and Yin Song. 2018.
\newblock \href {https://www.aclweb.org/anthology/C18-1066} {A position-aware
  bidirectional attention network for aspect-level sentiment analysis}.
\newblock In \emph{Proceedings of the 27th International Conference on
  Computational Linguistics}, pages 774--784, Santa Fe, New Mexico, USA.
  Association for Computational Linguistics.

\bibitem[{He et~al.(2019)He, Lee, Ng, and Dahlmeier}]{he2019interactive}
Ruidan He, Wee~Sun Lee, Hwee~Tou Ng, and Daniel Dahlmeier. 2019.
\newblock An interactive multi-task learning network for end-to-end
  aspect-based sentiment analysis.
\newblock In \emph{Proceedings of the 57th Annual Meeting of the Association
  for Computational Linguistics}, pages 504--515.

\bibitem[{Hu and Liu(2004)}]{10.1145/1014052.1014073}
Minqing Hu and Bing Liu. 2004.
\newblock \href {https://doi.org/10.1145/1014052.1014073} {Mining and
  summarizing customer reviews}.
\newblock In \emph{Proceedings of the Tenth ACM SIGKDD International Conference
  on Knowledge Discovery and Data Mining}, KDD '04, page 168–177, New York,
  NY, USA. Association for Computing Machinery.

\bibitem[{Jiang et~al.(2021)Jiang, Wang, and
  Aizawa}]{jiang-etal-2021-attention}
Junfeng Jiang, An~Wang, and Akiko Aizawa. 2021.
\newblock \href {https://www.aclweb.org/anthology/2021.eacl-main.170}
  {Attention-based relational graph convolutional network for target-oriented
  opinion words extraction}.
\newblock In \emph{Proceedings of the 16th Conference of the European Chapter
  of the Association for Computational Linguistics: Main Volume}, pages
  1986--1997, Online. Association for Computational Linguistics.

\bibitem[{Jiang et~al.(2019)Jiang, Chen, Xu, Ao, and
  Yang}]{jiang-etal-2019-challenge}
Qingnan Jiang, Lei Chen, Ruifeng Xu, Xiang Ao, and Min Yang. 2019.
\newblock \href {https://doi.org/10.18653/v1/D19-1654} {A challenge dataset and
  effective models for aspect-based sentiment analysis}.
\newblock In \emph{Proceedings of the 2019 Conference on Empirical Methods in
  Natural Language Processing and the 9th International Joint Conference on
  Natural Language Processing (EMNLP-IJCNLP)}, pages 6280--6285, Hong Kong,
  China. Association for Computational Linguistics.

\bibitem[{Kingma and Ba(2014)}]{kingma2014adam}
Diederik~P Kingma and Jimmy Ba. 2014.
\newblock Adam: A method for stochastic optimization.
\newblock \emph{arXiv preprint arXiv:1412.6980}.

\bibitem[{Li et~al.(2020{\natexlab{a}})Li, Chen, Quan, Ling, and
  Song}]{li-etal-2020-conditional}
Kun Li, Chengbo Chen, Xiaojun Quan, Qing Ling, and Yan Song.
  2020{\natexlab{a}}.
\newblock \href {https://doi.org/10.18653/v1/2020.acl-main.631} {Conditional
  augmentation for aspect term extraction via masked sequence-to-sequence
  generation}.
\newblock In \emph{Proceedings of the 58th Annual Meeting of the Association
  for Computational Linguistics}, pages 7056--7066, Online. Association for
  Computational Linguistics.

\bibitem[{Li et~al.(2018{\natexlab{a}})Li, Liu, and
  Zhou}]{li-etal-2018-hierarchical}
Lishuang Li, Yang Liu, and AnQiao Zhou. 2018{\natexlab{a}}.
\newblock \href {https://doi.org/10.18653/v1/K18-1018} {Hierarchical attention
  based position-aware network for aspect-level sentiment analysis}.
\newblock In \emph{Proceedings of the 22nd Conference on Computational Natural
  Language Learning}, pages 181--189, Brussels, Belgium. Association for
  Computational Linguistics.

\bibitem[{Li et~al.(2019)Li, Bing, Li, and Lam}]{li2019unified}
Xin Li, Lidong Bing, Piji Li, and Wai Lam. 2019.
\newblock A unified model for opinion target extraction and target sentiment
  prediction.
\newblock In \emph{Proceedings of the AAAI Conference on Artificial
  Intelligence}, volume~33, pages 6714--6721.

\bibitem[{Li et~al.(2018{\natexlab{b}})Li, Bing, Li, Lam, and
  Yang}]{li2018aspect}
Xin Li, Lidong Bing, Piji Li, Wai Lam, and Zhimou Yang. 2018{\natexlab{b}}.
\newblock Aspect term extraction with history attention and selective
  transformation.
\newblock \emph{arXiv preprint arXiv:1805.00760}.

\bibitem[{Li et~al.(2020{\natexlab{b}})Li, Yin, Zhong, and
  Pan}]{li-etal-2020-multi-instance}
Yuncong Li, Cunxiang Yin, Sheng-hua Zhong, and Xu~Pan. 2020{\natexlab{b}}.
\newblock \href {https://doi.org/10.18653/v1/2020.emnlp-main.287}
  {Multi-instance multi-label learning networks for aspect-category sentiment
  analysis}.
\newblock In \emph{Proceedings of the 2020 Conference on Empirical Methods in
  Natural Language Processing (EMNLP)}, pages 3550--3560, Online. Association
  for Computational Linguistics.

\bibitem[{Liu(2012)}]{liu2012sentiment}
Bing Liu. 2012.
\newblock Sentiment analysis and opinion mining.
\newblock \emph{Synthesis lectures on human language technologies},
  5(1):1--167.

\bibitem[{Ma et~al.(2017)Ma, Li, Zhang, and Wang}]{ma2017interactive}
Dehong Ma, Sujian Li, Xiaodong Zhang, and Houfeng Wang. 2017.
\newblock Interactive attention networks for aspect-level sentiment
  classification.
\newblock \emph{arXiv preprint arXiv:1709.00893}.

\bibitem[{Mao et~al.(2021)Mao, Shen, Yu, and Cai}]{mao2021joint}
Yue Mao, Yi~Shen, Chao Yu, and Longjun Cai. 2021.
\newblock A joint training dual-mrc framework for aspect based sentiment
  analysis.
\newblock \emph{arXiv preprint arXiv:2101.00816}.

\bibitem[{Nasukawa and Yi(2003)}]{10.1145/945645.945658}
Tetsuya Nasukawa and Jeonghee Yi. 2003.
\newblock \href {https://doi.org/10.1145/945645.945658} {Sentiment analysis:
  Capturing favorability using natural language processing}.
\newblock In \emph{Proceedings of the 2nd International Conference on Knowledge
  Capture}, K-CAP '03, page 70–77, New York, NY, USA. Association for
  Computing Machinery.

\bibitem[{Nguyen and Shirai(2015)}]{nguyen-shirai-2015-phrasernn}
Thien~Hai Nguyen and Kiyoaki Shirai. 2015.
\newblock \href {https://doi.org/10.18653/v1/D15-1298} {{P}hrase{RNN}: Phrase
  recursive neural network for aspect-based sentiment analysis}.
\newblock In \emph{Proceedings of the 2015 Conference on Empirical Methods in
  Natural Language Processing}, pages 2509--2514, Lisbon, Portugal. Association
  for Computational Linguistics.

\bibitem[{Paszke et~al.(2017)Paszke, Gross, Chintala, Chanan, Yang, DeVito,
  Lin, Desmaison, Antiga, and Lerer}]{paszke2017automatic}
Adam Paszke, Sam Gross, Soumith Chintala, Gregory Chanan, Edward Yang, Zachary
  DeVito, Zeming Lin, Alban Desmaison, Luca Antiga, and Adam Lerer. 2017.
\newblock Automatic differentiation in pytorch.

\bibitem[{Peng et~al.(2020)Peng, Xu, Bing, Huang, Lu, and
  Si}]{Peng2020KnowingWH}
H.~Peng, Lu~Xu, Lidong Bing, Fei Huang, Wei Lu, and L.~Si. 2020.
\newblock Knowing what, how and why: A near complete solution for aspect-based
  sentiment analysis.
\newblock In \emph{AAAI}.

\bibitem[{Phan and Ogunbona(2020)}]{phan2020modelling}
Minh~Hieu Phan and Philip~O Ogunbona. 2020.
\newblock Modelling context and syntactical features for aspect-based sentiment
  analysis.
\newblock In \emph{Proceedings of the 58th Annual Meeting of the Association
  for Computational Linguistics}, pages 3211--3220.

\bibitem[{Pontiki et~al.(2016)Pontiki, Galanis, Papageorgiou, Androutsopoulos,
  Manandhar, AL-Smadi, Al-Ayyoub, Zhao, Qin, De~Clercq, Hoste, Apidianaki,
  Tannier, Loukachevitch, Kotelnikov, Bel, Jim{\'e}nez-Zafra, and
  Eryi{\u{g}}it}]{pontiki-etal-2016-semeval}
Maria Pontiki, Dimitris Galanis, Haris Papageorgiou, Ion Androutsopoulos,
  Suresh Manandhar, Mohammad AL-Smadi, Mahmoud Al-Ayyoub, Yanyan Zhao, Bing
  Qin, Orph{\'e}e De~Clercq, V{\'e}ronique Hoste, Marianna Apidianaki, Xavier
  Tannier, Natalia Loukachevitch, Evgeniy Kotelnikov, Nuria Bel,
  Salud~Mar{\'\i}a Jim{\'e}nez-Zafra, and G{\"u}l{\c{s}}en Eryi{\u{g}}it. 2016.
\newblock \href {https://doi.org/10.18653/v1/S16-1002} {{S}em{E}val-2016 task
  5: Aspect based sentiment analysis}.
\newblock In \emph{Proceedings of the 10th International Workshop on Semantic
  Evaluation ({S}em{E}val-2016)}, pages 19--30, San Diego, California.
  Association for Computational Linguistics.

\bibitem[{Pontiki et~al.(2015)Pontiki, Galanis, Papageorgiou, Manandhar, and
  Androutsopoulos}]{pontiki-etal-2015-semeval}
Maria Pontiki, Dimitris Galanis, Haris Papageorgiou, Suresh Manandhar, and Ion
  Androutsopoulos. 2015.
\newblock \href {https://doi.org/10.18653/v1/S15-2082} {{S}em{E}val-2015 task
  12: Aspect based sentiment analysis}.
\newblock In \emph{Proceedings of the 9th International Workshop on Semantic
  Evaluation ({S}em{E}val 2015)}, pages 486--495, Denver, Colorado. Association
  for Computational Linguistics.

\bibitem[{Pontiki et~al.(2014)Pontiki, Galanis, Pavlopoulos, Papageorgiou,
  Androutsopoulos, and Manandhar}]{pontiki-etal-2014-semeval}
Maria Pontiki, Dimitris Galanis, John Pavlopoulos, Harris Papageorgiou, Ion
  Androutsopoulos, and Suresh Manandhar. 2014.
\newblock \href {https://doi.org/10.3115/v1/S14-2004} {{S}em{E}val-2014 task 4:
  Aspect based sentiment analysis}.
\newblock In \emph{Proceedings of the 8th International Workshop on Semantic
  Evaluation ({S}em{E}val 2014)}, pages 27--35, Dublin, Ireland. Association
  for Computational Linguistics.

\bibitem[{Pouran Ben~Veyseh et~al.(2020)Pouran Ben~Veyseh, Nouri, Dernoncourt,
  Dou, and Nguyen}]{pouran-ben-veyseh-etal-2020-introducing}
Amir Pouran Ben~Veyseh, Nasim Nouri, Franck Dernoncourt, Dejing Dou, and
  Thien~Huu Nguyen. 2020.
\newblock \href {https://doi.org/10.18653/v1/2020.emnlp-main.719} {Introducing
  syntactic structures into target opinion word extraction with deep learning}.
\newblock In \emph{Proceedings of the 2020 Conference on Empirical Methods in
  Natural Language Processing (EMNLP)}, pages 8947--8956, Online. Association
  for Computational Linguistics.

\bibitem[{Qiu et~al.(2011)Qiu, Liu, Bu, and Chen}]{qiu2011opinion}
Guang Qiu, Bing Liu, Jiajun Bu, and Chun Chen. 2011.
\newblock Opinion word expansion and target extraction through double
  propagation.
\newblock \emph{Computational linguistics}, 37(1):9--27.

\bibitem[{Stenetorp et~al.(2012)Stenetorp, Pyysalo, Topi{\'c}, Ohta, Ananiadou,
  and Tsujii}]{stenetorp-etal-2012-brat}
Pontus Stenetorp, Sampo Pyysalo, Goran Topi{\'c}, Tomoko Ohta, Sophia
  Ananiadou, and Jun{'}ichi Tsujii. 2012.
\newblock \href {https://www.aclweb.org/anthology/E12-2021} {brat: a web-based
  tool for {NLP}-assisted text annotation}.
\newblock In \emph{Proceedings of the Demonstrations at the 13th Conference of
  the {E}uropean Chapter of the Association for Computational Linguistics},
  pages 102--107, Avignon, France. Association for Computational Linguistics.

\bibitem[{Sutherland et~al.(2020)Sutherland, Bensch, Hellstr{\"o}m, Magg, and
  Wermter}]{10.1007/978-3-030-61609-0_52}
Alexander Sutherland, Suna Bensch, Thomas Hellstr{\"o}m, Sven Magg, and Stefan
  Wermter. 2020.
\newblock Tell me why you feel that way: Processing compositional dependency
  for tree-lstm aspect sentiment triplet extraction (taste).
\newblock In \emph{Artificial Neural Networks and Machine Learning -- ICANN
  2020}, pages 660--671, Cham. Springer International Publishing.

\bibitem[{Tang et~al.(2016)Tang, Qin, Feng, and Liu}]{tang-etal-2016-effective}
Duyu Tang, Bing Qin, Xiaocheng Feng, and Ting Liu. 2016.
\newblock \href {https://www.aclweb.org/anthology/C16-1311} {Effective {LSTM}s
  for target-dependent sentiment classification}.
\newblock In \emph{Proceedings of {COLING} 2016, the 26th International
  Conference on Computational Linguistics: Technical Papers}, pages 3298--3307,
  Osaka, Japan. The COLING 2016 Organizing Committee.

\bibitem[{Tang et~al.(2020)Tang, Ji, Li, and Zhou}]{tang2020dependency}
Hao Tang, Donghong Ji, Chenliang Li, and Qiji Zhou. 2020.
\newblock Dependency graph enhanced dual-transformer structure for aspect-based
  sentiment classification.
\newblock In \emph{Proceedings of the 58th Annual Meeting of the Association
  for Computational Linguistics}, pages 6578--6588.

\bibitem[{Wang et~al.(2020)Wang, Shen, Yang, Quan, and
  Wang}]{wang2020relational}
Kai Wang, Weizhou Shen, Yunyi Yang, Xiaojun Quan, and Rui Wang. 2020.
\newblock Relational graph attention network for aspect-based sentiment
  analysis.
\newblock \emph{arXiv preprint arXiv:2004.12362}.

\bibitem[{Wang et~al.(2016)Wang, Pan, Dahlmeier, and Xiao}]{wang2016recursive}
Wenya Wang, Sinno~Jialin Pan, Daniel Dahlmeier, and Xiaokui Xiao. 2016.
\newblock Recursive neural conditional random fields for aspect-based sentiment
  analysis.
\newblock \emph{arXiv preprint arXiv:1603.06679}.

\bibitem[{Wang et~al.(2017)Wang, Pan, Dahlmeier, and Xiao}]{wang2017coupled}
Wenya Wang, Sinno~Jialin Pan, Daniel Dahlmeier, and Xiaokui Xiao. 2017.
\newblock Coupled multi-layer attentions for co-extraction of aspect and
  opinion terms.
\newblock In \emph{Thirty-First AAAI Conference on Artificial Intelligence}.

\bibitem[{Wei et~al.(2020)Wei, Hong, Zou, Cheng, and Jianmin}]{wei2020don}
Zhenkai Wei, Yu~Hong, Bowei Zou, Meng Cheng, and YAO Jianmin. 2020.
\newblock Don’t eclipse your arts due to small discrepancies: Boundary
  repositioning with a pointer network for aspect extraction.
\newblock In \emph{Proceedings of the 58th Annual Meeting of the Association
  for Computational Linguistics}, pages 3678--3684.

\bibitem[{Wu and He(2019)}]{wu2019enriching}
Shanchan Wu and Yifan He. 2019.
\newblock Enriching pre-trained language model with entity information for
  relation classification.
\newblock In \emph{Proceedings of the 28th ACM International Conference on
  Information and Knowledge Management}, pages 2361--2364.

\bibitem[{Wu et~al.(2020{\natexlab{a}})Wu, Ying, Zhao, Fan, Dai, and
  Xia}]{wu-etal-2020-grid}
Zhen Wu, Chengcan Ying, Fei Zhao, Zhifang Fan, Xinyu Dai, and Rui Xia.
  2020{\natexlab{a}}.
\newblock \href {https://doi.org/10.18653/v1/2020.findings-emnlp.234} {Grid
  tagging scheme for aspect-oriented fine-grained opinion extraction}.
\newblock In \emph{Findings of the Association for Computational Linguistics:
  EMNLP 2020}, pages 2576--2585, Online. Association for Computational
  Linguistics.

\bibitem[{Wu et~al.(2020{\natexlab{b}})Wu, Zhao, Dai, Huang, and
  Chen}]{wu2020latent}
Zhen Wu, Fei Zhao, Xin-Yu Dai, Shujian Huang, and Jiajun Chen.
  2020{\natexlab{b}}.
\newblock Latent opinions transfer network for target-oriented opinion words
  extraction.
\newblock \emph{arXiv preprint arXiv:2001.01989}.

\bibitem[{Xu et~al.(2018)Xu, Liu, Shu, and Philip}]{xu2018double}
Hu~Xu, Bing Liu, Lei Shu, and S~Yu Philip. 2018.
\newblock Double embeddings and cnn-based sequence labeling for aspect
  extraction.
\newblock In \emph{Proceedings of the 56th Annual Meeting of the Association
  for Computational Linguistics (Volume 2: Short Papers)}, pages 592--598.

\bibitem[{Xu et~al.(2020)Xu, Li, Lu, and Bing}]{xu-etal-2020-position}
Lu~Xu, Hao Li, Wei Lu, and Lidong Bing. 2020.
\newblock \href {https://doi.org/10.18653/v1/2020.emnlp-main.183}
  {Position-aware tagging for aspect sentiment triplet extraction}.
\newblock In \emph{Proceedings of the 2020 Conference on Empirical Methods in
  Natural Language Processing (EMNLP)}, pages 2339--2349, Online. Association
  for Computational Linguistics.

\bibitem[{Xue and Li(2018)}]{xue-li-2018-aspect}
Wei Xue and Tao Li. 2018.
\newblock \href {https://doi.org/10.18653/v1/P18-1234} {Aspect based sentiment
  analysis with gated convolutional networks}.
\newblock In \emph{Proceedings of the 56th Annual Meeting of the Association
  for Computational Linguistics (Volume 1: Long Papers)}, pages 2514--2523,
  Melbourne, Australia. Association for Computational Linguistics.

\bibitem[{Yin et~al.(2016)Yin, Wei, Dong, Xu, Zhang, and
  Zhou}]{yin2016unsupervised}
Yichun Yin, Furu Wei, Li~Dong, Kaimeng Xu, Ming Zhang, and Ming Zhou. 2016.
\newblock Unsupervised word and dependency path embeddings for aspect term
  extraction.
\newblock \emph{arXiv preprint arXiv:1605.07843}.

\bibitem[{Zhang et~al.(2020)Zhang, Li, Song, and
  Wang}]{zhang-etal-2020-multi-task}
Chen Zhang, Qiuchi Li, Dawei Song, and Benyou Wang. 2020.
\newblock \href {https://doi.org/10.18653/v1/2020.findings-emnlp.72} {A
  multi-task learning framework for opinion triplet extraction}.
\newblock In \emph{Findings of the Association for Computational Linguistics:
  EMNLP 2020}, pages 819--828, Online. Association for Computational
  Linguistics.

\bibitem[{Zhao et~al.(2020{\natexlab{a}})Zhao, Huang, Zhang, Lu
  et~al.}]{zhao2020spanmlt}
He~Zhao, Longtao Huang, Rong Zhang, Quan Lu, et~al. 2020{\natexlab{a}}.
\newblock Spanmlt: A span-based multi-task learning framework for pair-wise
  aspect and opinion terms extraction.
\newblock In \emph{Proceedings of the 58th Annual Meeting of the Association
  for Computational Linguistics}, pages 3239--3248.

\bibitem[{Zhao et~al.(2020{\natexlab{b}})Zhao, Hou, and Wu}]{zhao2020modeling}
Pinlong Zhao, Linlin Hou, and Ou~Wu. 2020{\natexlab{b}}.
\newblock Modeling sentiment dependencies with graph convolutional networks for
  aspect-level sentiment classification.
\newblock \emph{Knowledge-Based Systems}, 193:105443.

\bibitem[{Zhong and Chen(2020)}]{zhong2020frustratingly}
Zexuan Zhong and Danqi Chen. 2020.
\newblock A frustratingly easy approach for joint entity and relation
  extraction.
\newblock \emph{arXiv preprint arXiv:2010.12812}.

\bibitem[{Zhou et~al.(2019)Zhou, Huang, Guo, Han, and Hu}]{zhou2019span}
Yan Zhou, Longtao Huang, Tao Guo, Jizhong Han, and Songlin Hu. 2019.
\newblock A span-based joint model for opinion target extraction and target
  sentiment classification.
\newblock In \emph{IJCAI}, pages 5485--5491.

\end{thebibliography}
\bibliographystyle{acl_natbib}

\appendix

\section{More Related Work}
\label{sec:More_Related_Work}
\citet{wang2016recursive, wang2017coupled} have annotated the opinions and thier sentiments of the sentences in the restaurant and laptop datasets from SemEval-2014 Task 4\citep{pontiki-etal-2014-semeval} and the restaurant dataset from SemEval-2015 Task 12\citep{pontiki-etal-2015-semeval}. Is it necessary to annotate the sentiments of the aspect and opinion pairs in the Aspect Sentiment Triplet Extraction (\textbf{ASTE}) datasets for obtaining our Aspect-Sentiment-Opinion Triplet Extraction (\textbf{ASOTE}) datasets? The answer is yes. The reasons are as follows:
\begin{itemize}
	\item The sentiments of aspect and opinion pairs are different from the sentiments of opinions.
	\item The opinions annotated by \citet{wang2016recursive, wang2017coupled} are different from the opinions annotated in the Target-oriented Opinion Words Extraction (\textbf{TOWE}) datasets \citep{fan2019target} which are used to construct our ASOTE datasets. For example, given this sentence, "those rolls were big , but not good and sashimi wasn't fresh.", the opinions and their sentiments annotated by \citet{wang2016recursive, wang2017coupled} are (``big'', positive), (``good'', positive), and (``fresh'', positive), while the opinions annotated in the TOWE datasets are "big", "not good" and "wasn't fresh" and the triplets containing "not good" and "wasn't fresh" are negative. We think the opinions annotated in the TOWE datasets are more appropriate for the ASOTE task.
\end{itemize}

\citet{zhang-etal-2020-multi-task}  defined their Opinion Triplet Extraction task as an integration of aspect-sentiment pair extraction~\citep{zhou2019span, li2019unified, phan2020modelling} and aspect-opinion co-extraction~\citep{wang2016recursive, wang2017coupled, dai2019neural}. The obtained Opinion Triplet Extraction task has the same goal as our ASOTE task. However, the authors used the ASTE datasets to evaluate the perfromances of their models on the Opinion Triplet Extraction task and said that the Opinion Triplet Extraction task is the same as the ASTE task in the corresponding github repository of their paper. In fact, combining aspect-sentiment pair extraction with aspect-opinion co-extraction can obtain neither the ASTE task nor our ASOTE task. Therefore, we don't think they introduce the ASOTE task.

\section{More Data Statistics}
More statistics about our ASOTE-data are shown in Table~\ref{table:more-dataset-statistics}.

\begin{table*}
	\centering
	\begin{tabular}{|l|l|l|l|}
		\hline
		\multicolumn{2}{|l|}{Dataset}  & \#aspect\_with\_multiple\_opinions & \#opinion\_with\_multiple\_aspects \\ \hline
		\multirow{3}{*}{14res} & train & 305                                & 211                                \\ \cline{2-4} 
		& dev   & 54                                 & 48                                 \\ \cline{2-4} 
		& test  & 143                                & 88                                 \\ \hline
		\multirow{3}{*}{14lap} & train & 179                                & 157                                \\ \cline{2-4} 
		& dev   & 37                                 & 35                                 \\ \cline{2-4} 
		& test  & 70                                 & 53                                 \\ \hline
		\multirow{3}{*}{15res} & train & 143                                & 45                                 \\ \cline{2-4} 
		& dev   & 30                                 & 15                                 \\ \cline{2-4} 
		& test  & 51                                 & 23                                 \\ \hline
		\multirow{3}{*}{16res} & train & 186                                & 66                                 \\ \cline{2-4} 
		& dev   & 37                                 & 18                                 \\ \cline{2-4} 
		& test  & 61                                 & 23                                 \\ \hline
	\end{tabular}
	\caption{\label{table:more-dataset-statistics} More statistics.}
\end{table*}

\section{More Experimental Results}
\label{sec:Experimental_Results}

\begin{table*}
	\centering
	\begin{tabular}{|l|l|l|l|l|l|l|l|l|l|l|l|l|}
		\hline
		& \multicolumn{3}{l|}{14res} & \multicolumn{3}{l|}{14lap}  & \multicolumn{3}{l|}{15res}  & \multicolumn{3}{l|}{16res}  \\ \hline
		Method      & P     & R    & F1          & P    & R    & F1            & P    & R    & F1            & P    & R    & F1            \\ \hline
		GTS-CNN     & 64.2  & 61.7 & 62.9        & 56.4 & 44.1 & 49.4          & 54.4 & 54.4 & 54.1          & 59.5 & 63.9 & 61.4          \\ \hline
		GTS-BiLSTM  & 71.6  & 58.7 & 64.5        & 67.3 & 38.1 & 48.5          & 69.0   & 48.4 & 56.8          & 66.8 & 59.4 & 62.6          \\ \hline
		GTS-BERT    & 72.3  & 71.1 & 71.7        & 64.5 & 56.6 & 60.2          & 63.8 & 59.3 & 61.5          & 64.9 & 71.9 & 68.1          \\ \hline
		PBF         & 74.2  & 73.9 & \textbf{74.0} & 64.7 & 63.0   & 63.8          & 61.0   & 67.3 & 63.9          & 65.2 & 77.4 & \textbf{70.8} \\ \hline
		PBF -w/o A  & 72.1  & 74.3 & 73.2        & 64.0   & 63.8 & 63.9          & 61.8 & 67.5 & \textbf{64.5} & 65.4 & 76.9 & 70.6          \\ \hline
		PBF -w/o P  & 73.0    & 74.2 & 73.6        & 65.3 & 63.2 & \textbf{64.3} & 61.2 & 65.8 & 63.3          & 64.0   & 77.1 & 69.9          \\ \hline
		PBF -w/o AP & 48.0    & 56.1 & 51.3        & 52.7 & 57.1 & 54.6          & 45.1 & 55.0   & 49.4          & 49.9 & 64.7 & 56.3          \\ \hline
		PBF-M1      & 71.2  & 74.5 & 72.8        & 66.9 & 61.6 & 64.0            & 63.4 & 65.1 & 64.1          & 63.7 & 77.5 & 69.9          \\ \hline
		PBF-M2      & 68.1  & 68.8 & 68.4        & 60.0   & 54.8 & 57.3          & 56.3 & 63.0   & 59.4          & 61.6 & 71.7 & 66.2          \\ \hline
		PBF-M3      & 71.5  & 74.2 & 72.8        & 65.8 & 61.7 & 63.7          & 59.4 & 67.7 & 63.2          & 64.5 & 76.0   & 69.7          \\ \hline
	\end{tabular}
	\caption{\label{table:AOPE-more} Results of aspect-opinion pair extraction on the ASOTE-data.}
\end{table*}

\begin{table*}
	\centering
	\begin{tabular}{|l|l|l|l|l|l|l|l|l|l|l|l|l|}
		\hline
		& \multicolumn{3}{l|}{14res}  & \multicolumn{3}{l|}{14lap}  & \multicolumn{3}{l|}{15res}  & \multicolumn{3}{l|}{16res}  \\ \hline
		Method      & P    & R    & F1            & P    & R    & F1            & P    & R    & F1            & P    & R    & F1            \\ \hline
		PBF         & 82.8 & 80.3 & \textbf{81.5} & 73.6 & 74.5 & 74.0            & 75.5 & 80.4 & 77.9          & 77.1 & 87.9 & \textbf{82.1} \\ \hline
		PBF -w/o A  & 80.5 & 80.9 & 80.7          & 72.7 & 75.7 & \textbf{74.1} & 76.9 & 80.4 & \textbf{78.6} & 76.8 & 87.2 & 81.6          \\ \hline
		PBF -w/o P  & 81.3 & 80.5 & 80.9          & 73.8 & 74.1 & 74.0            & 76.2 & 78.6 & 77.3          & 75.2 & 87.7 & 81.0            \\ \hline
		PBF -w/o AP & 53.0   & 60.8 & 56.1          & 58.9 & 66.0   & 61.9          & 56.2 & 65.7 & 60.5          & 58.0   & 73.7 & 64.8          \\ \hline
		PBF-M1      & 79.4 & 80.8 & 80.1          & 74.8 & 71.4 & 73.0            & 78.1 & 76.8 & 77.4          & 74.0   & 88.3 & 80.5          \\ \hline
		PBF-M2      & 75.6 & 74.6 & 75.1          & 67.5 & 64.9 & 66.1          & 71.4 & 74.5 & 72.9          & 72.2 & 81.5 & 76.5          \\ \hline
		PBF-M3      & 80.0   & 80.8 & 80.3          & 74.0   & 73.1 & 73.6          & 74.6 & 80.7 & 77.5          & 75.6 & 85.8 & 80.3          \\ \hline
	\end{tabular}
	\caption{\label{table:TOWE-ASOTE-data-prf} Results of the TOWE task on the ASOTE-data. }
\end{table*}

\begin{table*}
	\centering
	\begin{tabular}{|l|l|l|l|l|l|l|l|l|l|l|l|l|}
		\hline
		& \multicolumn{3}{l|}{14res}  & \multicolumn{3}{l|}{14lap}  & \multicolumn{3}{l|}{15res}  & \multicolumn{3}{l|}{16res}  \\ \hline
		Method          & P    & R    & F1            & P    & R    & F1            & P    & R    & F1            & P    & R    & F1            \\ \hline
		IOG             & 82.8 & 77.3 & 80.0            & 73.2 & 69.6 & 71.3          & 76.0   & 70.7 & 73.2          & 85.2 & 78.5 & 81.6          \\ \hline
		LOTN            & 84.0   & 80.5 & 82.2          & 77.0   & 67.6 & 72.0            & 76.6 & 70.2 & 73.2          & 86.5 & 80.8 & 83.6          \\ \hline
		ARGCN           & 86.6 & 82.7 & 84.6          & 79.4 & 71.6 & 75.3          & 76.5 & 76.8 & 76.7          & 86.1 & 84.1 & 85.1          \\ \hline
		ARGCN$_{+bert}$ & 87.3 & 83.5 & 85.4          & 75.8 & 76.9 & 76.3          & 78.8 & 77.6 & 78.2          & 88.4 & 84.9 & 86.6          \\ \hline
		ONG             & 83.2 & 81.4 & 82.3          & 73.8 & 77.7 & 75.7          & 76.6 & 81.1 & 78.8          & 87.7 & 84.3 & 86.0            \\ \hline
		PBF             & 85.5 & 86.2 & 85.9          & 81.8 & 81.2 & \textbf{81.5} & 79.3 & 82.4 & \textbf{80.8} & 89.2 & 89.3 & \textbf{89.2} \\ \hline
		PBF -w/o A      & 85.9 & 86.3 & 86.1          & 82.4 & 80.0   & 81.2          & 78.4 & 82.5 & 80.4          & 86.8 & 89.1 & 87.9          \\ \hline
		PBF -w/o P      & 85.7 & 87.0   & \textbf{86.3} & 80.2 & 80.4 & 80.3          & 77.5 & 82.4 & 79.8          & 87.4 & 90.3 & 88.8          \\ \hline
		PBF -w/o AP     & 57.9 & 66.2 & 61.6          & 63.1 & 73.8 & 67.9          & 50.4 & 71.8 & 59.0            & 65.2 & 74.0   & 69.3          \\ \hline
	\end{tabular}
	\caption{\label{table:TOWE-TOWE-data-prf} Results of the TOWE task on the TOWE-data.}
\end{table*}

The results including precision (P), recall (R) and F1 (F) score of the aspect-opinion pair extraction task  on the ASOTE-data are shown in Table~\ref{table:AOPE-more}.

The results including precision (P), recall (R) and F1 (F) score of the TOWE task  on the ASOTE-data are shown in Table~\ref{table:TOWE-ASOTE-data-prf}.

The results including precision (P), recall (R) and F1 (F) score of the TOWE task  on the TOWE-data are shown in Table~\ref{table:TOWE-TOWE-data-prf}.

The results of the Aspect-Opinion Pair Sentiment Classification (AOPSC) task are shown in Table~\ref{table:AOPSC}. Position-aware BERT-based Framework (PBF) and its variants including PBF -w/o AP obtain similar performance. 

\begin{table}
	\centering
	\begin{tabular}{|l|l|l|l|l|}
		\hline
		Method        & 14res       & 14lap         & 15res         & 16res         \\ \hline
		PBF -w/o AP   & 90.8        & 82.9          & 89.2          & 91.2          \\ \hline
		PBF           & 90.8        & \textbf{83.4} & 88.5          & 92.0            \\ \hline
		PBF -w/o P    & \textbf{91.0} & 82.8          & 88.8          & 91.6          \\ \hline
		PBF -w/o A & 90.6        & 83.2          & \textbf{89.3} & 91.5          \\ \hline
		PBF-M1     & 90.9        & 83.2          & 88.2          & 91.5          \\ \hline
		PBF-M2     & 89.2        & 80.7          & 85.2          & 89.2          \\ \hline
		PBF-M3        & 90.9        & 82.6          & 89.0            & \textbf{92.2} \\ \hline
	\end{tabular}
	\caption{\label{table:AOPSC} Results of AOPSC task in terms of accuracy.}
\end{table}

The results of the Aspect Term Extraction (ATE) task are shown in Table~\ref{table:ATE}.
\begin{table*}
	\centering
	\begin{tabular}{|l|l|l|l|l|l|l|l|l|l|l|l|l|}
		\hline
		& \multicolumn{3}{l|}{14res} & \multicolumn{3}{l|}{14lap} & \multicolumn{3}{l|}{15res} & \multicolumn{3}{l|}{16res} \\ \hline
		Method & P      & R       & F1      & P        & R       & F1    & P        & R     & F1      & P       & R       & F1     \\ \hline
		PBF    & 87     & 88.4    & 87.7    & 82.7     & 81.3    & 82    & 68.4     & 73    & 70.5    & 75.3    & 76.5    & 75.9   \\ \hline
	\end{tabular}
	\caption{\label{table:ATE} Results of the Aspect Term Extraction (ATE) task.}
\end{table*}

\subsection{Case Study}
\begin{figure*}
	\centering
	\includegraphics[scale=0.25]{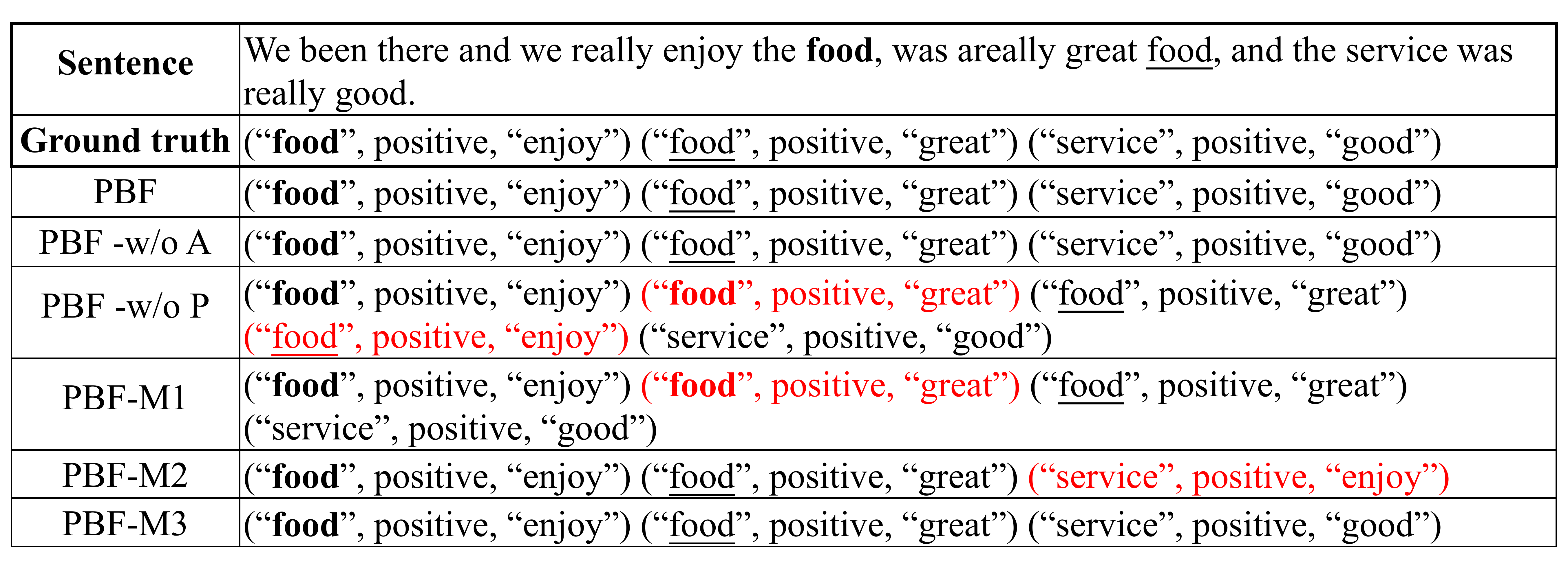}
	\caption{Case study. Red triplets are incorrect predictions.}
	\label{fig:more_case_study1}
\end{figure*}

The extraction results of our Position-aware BERT-based Framework (PBF) on the sentence ``We been there and we really enjoy the food, was areally great food, and the service was really good.
'' are shown in Figure~\ref{fig:more_case_study1}. From the results, we can see that all variants except for PBF -w/o P include the position information of aspects.

\begin{figure*}
	\centering
	\includegraphics[scale=0.22]{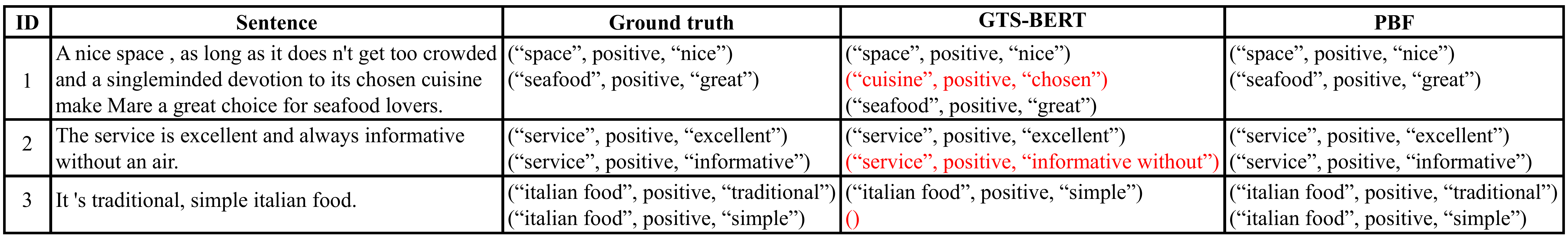}
	\caption{Case study. Red triplets are incorrect predictions.}
	\label{fig:more_case_study2}
\end{figure*}

Figure~\ref{fig:more_case_study2} shows three sentences. While PBF correctly extracts all triplets from these sentences, GTS-BERT can't correctly extract all triplets from these sentences.

Table~\ref{fig:more_case_study3} shows some hard sentences that both GTS-BERT and PBF can not correctly extract all triplets from.

\begin{table*}
	\centering
	\small
	\begin{tabular}{|l|l|l|l|}
		\hline
		Sentence                                                                                    & Ground Truth                                                                                                                                                                  & GTS-BERT                                                                                                                         & PBF                                                                                                                                                    \\ \hline
		\begin{tabular}[c]{@{}l@{}}Simple healthy \\ unglamorous \\ food cheap.\end{tabular}        & \begin{tabular}[c]{@{}l@{}}("food", positive, "Simple"),\\ ("food",   positive, "cheap"),\\ ("food", positive,   "healthy"),\\ ("food", positive, "unglamorous")\end{tabular} & \begin{tabular}[c]{@{}l@{}}("food", positive, \\ "Simple healthy \\ unglamorous")\end{tabular}                                   & \begin{tabular}[c]{@{}l@{}}("food", positive, \\ "cheap"),\\ ("food",   positive, \\ "healthy"),\\ ("food", positive,   \\ "unglamorous")\end{tabular}                                                                    \\ \hline
		\begin{tabular}[c]{@{}l@{}}The staff should be \\ a bit more friendly .\end{tabular}        & ("staff", negative, "friendly")                                                                                                                                               & ("staff", positive, "friendly")                                                                                                  & \begin{tabular}[c]{@{}l@{}}("staff", positive, \\ "friendly")\end{tabular}                                                                             \\ \hline
		\begin{tabular}[c]{@{}l@{}}The gnocchi literally \\ melts in your mouth!\end{tabular}       & ("gnocchi", positive, "melts")                                                                                                                                                &                                                                                                                                  &                                                                                                                                                                                                \\ \hline
		\begin{tabular}[c]{@{}l@{}}The avocado salad is \\ a personal fave .\end{tabular}           & \begin{tabular}[c]{@{}l@{}}("avocado salad", positive, \\ "fave")\end{tabular}                                                                                                &                                                                                                                                  &                                                                                                                                                        \\ \hline
		\begin{tabular}[c]{@{}l@{}}good food good \\ wine that 's it .\end{tabular}                 & \begin{tabular}[c]{@{}l@{}}("food", positive, "good"),\\ ("wine",   positive, "good")\end{tabular}                                                                            & \begin{tabular}[c]{@{}l@{}}("food", positive, "good"),\\ ("wine",   positive, "good"),\\ ("wine", positive, "good")\end{tabular} & \begin{tabular}[c]{@{}l@{}}("wine", positive, \\ "good")\end{tabular}                                                                                  \\ \hline
		\begin{tabular}[c]{@{}l@{}}Save room for deserts \\ - they 're to die   for .\end{tabular}  & ("deserts", positive, "die for")                                                                                                                                              &                                                                                                                                  &                                                                                                                                                                                                                                \\ \hline
		\begin{tabular}[c]{@{}l@{}}Service was good and \\ so was the   atmosphere.\end{tabular}    & \begin{tabular}[c]{@{}l@{}}("Service", positive,   "good"),\\ ("atmosphere", positive, "good")\end{tabular}                                                                   & \begin{tabular}[c]{@{}l@{}}("Service", positive, \\ "good")\end{tabular}                                                         & \begin{tabular}[c]{@{}l@{}}("Service", positive, \\ "good")\end{tabular}                                                                               \\ \hline
		\begin{tabular}[c]{@{}l@{}}Try the homemade \\ breads.\end{tabular}                         & \begin{tabular}[c]{@{}l@{}}("homemade breads", positive, \\ "Try")\end{tabular}                                                                                               & ("breads", positive, "Try")                                                                                                      &                                                                                                                                                        \\ \hline
		\begin{tabular}[c]{@{}l@{}}The help was extremely \\ nice and did not rush us.\end{tabular} &                                                                                                                                                                               & ("help", positive, "nice")                                                                                                       & \begin{tabular}[c]{@{}l@{}}("help", positive, \\ "nice")\end{tabular}                                                                                  \\ \hline
	\end{tabular}
	\caption{Case study. Some sentences that both GTS-BERT and PBF can not correctly extract all triplets from.}
	\label{fig:more_case_study3}
\end{table*}

\end{document}